\pdfoutput=1

\documentclass[11pt]{article}

\usepackage[final]{acl}

\usepackage{times}
\usepackage{latexsym}

\usepackage[T1]{fontenc}

\usepackage[utf8]{inputenc}

\usepackage{microtype}

\usepackage{inconsolata}

\usepackage{graphicx}
\usepackage{amssymb}
\usepackage{amsmath}
\usepackage{multirow}
\usepackage{booktabs}
\usepackage{colortbl}
\usepackage{makecell}
\usepackage{subfigure}

\title{Unveiling Linguistic Regions in Large Language Models}

\author{Zhihao Zhang$^1$\thanks{{ }\ Equal contributions.}, Jun Zhao$^1$$^{*}$, Qi Zhang$^1$$^3$\thanks{{ }\ Corresponding authors.},  Tao Gui$^2$, Xuanjing Huang$^1$\\
$^1$ School of Computer Science, Fudan University\\
$^2$ Institute of Modern Languages and Linguistics, Fudan University \\
$^3$ Shanghai Collaborative Innovation Center of Intelligent Visual Computing \\
\texttt{\{zhangzhihao19, zhaoj19, qz, tgui, xjhuang\}@fudan.edu.cn} \\
}

\begin{document}
\maketitle
\begin{abstract}
Large Language Models (LLMs) have demonstrated considerable cross-lingual alignment and generalization ability. Current research primarily focuses on improving LLMs' cross-lingual generalization capabilities. However, there is still a lack of research on the intrinsic mechanisms of how LLMs achieve cross-lingual alignment. From the perspective of region partitioning, this paper conducts several investigations on the linguistic competence of LLMs. We discover a core region in LLMs that corresponds to linguistic competence, accounting for approximately 1\% of the total model parameters. Removing this core region by setting parameters to zero results in a significant performance decrease across 30 different languages. Furthermore, this core region exhibits significant dimensional dependence, perturbations to even a single parameter on specific dimensions leading to a loss of linguistic competence. Moreover, we discover that distinct monolingual regions exist for different languages, and disruption to these specific regions substantially reduces the LLMs' proficiency in those corresponding languages. Our research also indicates that freezing the core linguistic region during further pre-training can mitigate the issue of catastrophic forgetting (CF), a common phenomenon observed during further pre-training of LLMs. Overall, exploring the LLMs' functional regions provides insights into the foundation of their intelligence \footnote{\hspace{-0.37em} Our code is released in \url{https://github.com/zzhang0179/Unveiling-Linguistic-Regions-in-LLMs}.}.
\end{abstract}

\section{Introduction}
%
\begin{figure}[t]
    \includegraphics[width=\linewidth]{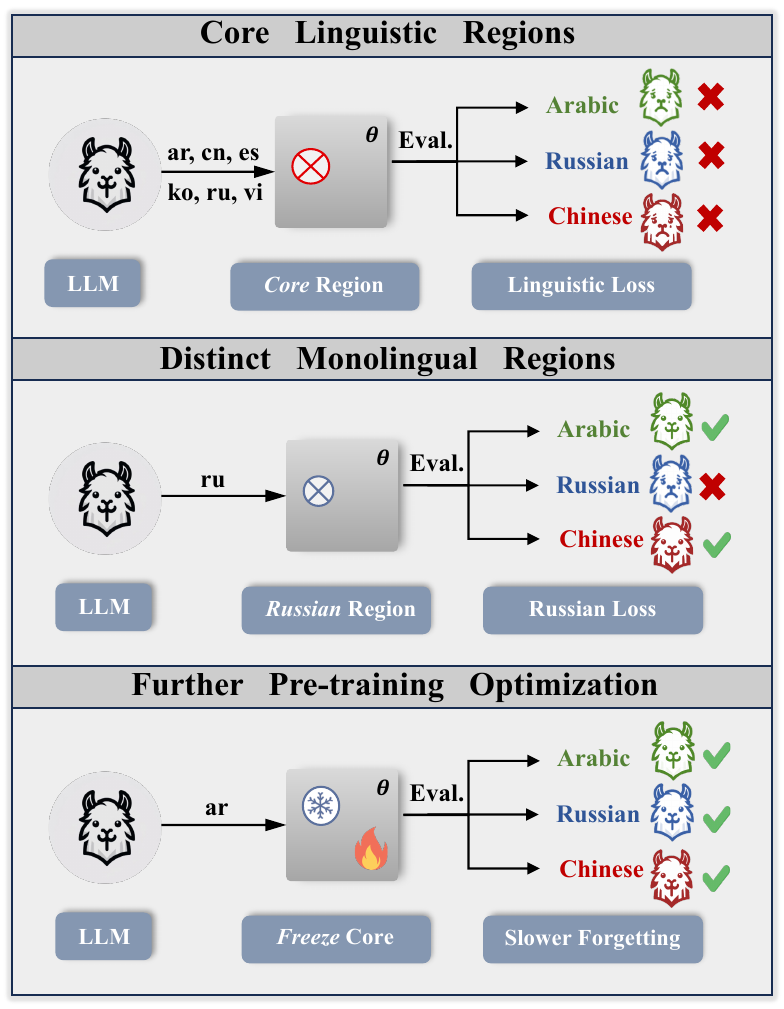}
    \caption{Three main findings of our experiments: (1) Identification of core language regions within the LLMs, where removals lead to linguistic competence loss; (2) Discovery of monolingual regions, where removals cause significant proficiency loss in specific languages; (3) Optimization of freezing core regions during further pre-training decelerates language forgetting.
    }
    \label{fig:introduction}
\end{figure}

Over the years, the field of Natural Language Processing (NLP) has been at the forefront of understanding the core principles of intelligence. The emergence of Large Language Models (LLMs) such as GPT-4 \citep{GPT-4}, PaLM 2 \citep{PaLM2} and LLaMA 2 \citep{LLaMA2}, showcase a significant breakthrough. 
Thanks to unparalleled scales of model architecture and the vastness of training data, these LLMs now exhibit exceptional linguistic competence and can execute complex tasks requiring abstract knowledge \citep{dong2023survey} and reasoning \citep{DBLP:journals/corr/abs-2110-14168}.

Previous research has revealed that LLMs naturally capture cross-linguistic similarities in their representation space, facilitating zero-shot cross-lingual transfer \citep{How-Multilingual-is-Multilingual-BERT,The-Surprising-Cross-Lingual-Effectiveness-of-BERT,Are-Structural-Concepts-Universal-in-Transformer-Language-Models}. The model is fine-tuned on one language, enabling the acquisition of comparable capabilities in another language \citep{Crosslingual-Generalization-through-Multitask-Finetuning,An-Empirical-Revisiting-on-Multilingual-Transfer-Ability}, and exhibits the phenomenon of code-switching when generating context
\citep{GLUECoS,LLaMA-Beyond-English}. Attempts to improve LLMs' cross-lingual generalization abilities have been successful through parameter and information transfer learning \citep{UDapter,Cross-Lingual-Transfer}, aligning languages compulsorily \citep{Zero-Shot-Cross-lingual-Semantic-Parsing,Multilingual-Instruction-Tuning-With-Just-a-Pinch-of-Multilinguality} and utilizing in-context learning techniques \citep{few-shot-mullearning,multi-icl-alignment}. However, a detailed investigation into the internal mechanisms of how LLMs possess cross-linguistic alignment capability remains elusive.

To delve deeper into the intrinsic mechanisms of LLMs' linguistic competence, this paper focuses on the LLMs' parameter importance and investigate the linguistic regions of LLMs based on 30 distinct languages' performance, with the purpose of figuring out the following questions:

\textbf{Q1: Does a core linguistic region exist within LLMs that facilitates cross-lingual alignment and generalization?}
By conducting further pre-training across six languages and evaluating models' parameter importance (Section \ref{sec:importance}), we discover a region in LLMs corresponding to the core linguistic competence, which accounts for approximately 1\% of the model's total parameters. As shown at the top of Figure \ref{fig:introduction}, removing this region (setting parameters to \textit{zero}) consistently leads to a significant decline in performance across 30 test languages (Section \ref{sec:localization}). 

Furthermore, by visualizing the core linguistic region (Figure \ref{fig:visualize}), we observe that the linguistic core region of LLMs exhibits significant dimensional dependence. In certain dimensions, only perturbing a single parameter could lead to the model losing its linguistic competence (Section \ref{sec:dependence}). Additionally, ablation study in \ref{paragraph:ablation_study} shows that beyond outlier dimensions, other non-outlier dimensions in this region are also critical.

\textbf{Q2: Beyond the core linguistic region within LLMs , do distinct monolingual regions exist that specifically influence individual languages?}
While LLMs possess strong multilingual capabilities, we discover that each individual language (or language with similar compositional elements or grammatical structures) encompasses independent regions within the LLMs. As shown in the middle of Figure \ref{fig:introduction}, the analysis of the Russian sentences identifies a particular linguistic region that likewise exerts influence both on the Russian and Ukrainian language, both of which belong to the Slavic group (Section \ref{sec:single localization}).

\textbf{Q3: If and how core linguistic regions affect further pre-training, how to utilize it to optimize further pre-training?} 
After pre-training, core linguistic parameter regions of the LLMs are established for multilingual alignment. Notable shifts in these regions potentially lead to a decline in model lingual capabilities.
Our findings reveal that freezing this core region can mitigate the issue of catastrophic forgetting \citep{CF-in-Connectionist-Networks, Measuring-CF}, a common phenomenon observed during further pre-training of LLMs. As shown at the bottom of Figure \ref{fig:introduction}, we investigate the impact of selectively freezing 5\% key parameters of all parameters during further pre-training, compared to the full-scale fine-tuning technique. Findings indicate that this method facilitates comparable learning of the target language while concurrently decelerating the rate of language attrition for previously learned languages (Section \ref{sec:Further Pre-training Optimization}). Significantly, our methodology is compatible with the data-replay techniques \citep{Catastrophic-Forgetting-Rehearsal-and-Pseudorehearsal,Skywork}, with no necessity for integrating extra components into the model. Unlike regularization methods \citep{Dropout,L2regularization}, our approach restricts to a minimal core region in LLMs.

The main contributions of our work are summarized as follows:
\begin{itemize}
  \item We discover that LLMs possess a core linguistic region, and removing this region (setting parameters to \textit{zero}) results in a significant loss of the model's linguistic capabilities. Furthermore, perturbations to specific dimensions or even a single parameter can lead to a substantial decline in the model's linguistic abilities.
  \item We observe that distinct monolingual regions exist in LLMs for different languages. Removing a specific monolingual region causes a significant deterioration in the linguistic capabilities within corresponding language.
  \item We perform further pre-training for specific languages within the core linguistic region of LLMs frozen, achieving comparable performance in the target language while mitigating catastrophic forgetting in non-target languages.
\end{itemize}

\section{Background and Metric}
\subsection{Model Pre-training}
\label{sec:pretrain}
Pre-training is a crucial process by which LLMs acquire linguistic competence and gain general knowledge about the real world. Formally, given a large corpus $\mathcal{D}$, the training objective for auto-regressive language modeling is to find the optimal $\theta$ that minimizes the following loss $\mathcal{L}$:
\begin{equation}
\mathcal{L}{(\mathcal{D},\theta)} =\sum_{x\in\mathcal{D}}\sum_i\log p_\theta(x_i|x_1,...,x_{i-1}),
\end{equation}
where $x=\{x_1, ..., x_n\}$ denotes an input token sequence and $\theta$ denotes parameters of the model.

\subsection{Parameter Importance}
\label{sec:importance}
Drawing upon the observations of linguistic alignments, we propose that particular parameters regions within the model exert significant influence on its inherent language alignment capabilities. Evaluating parameter sensitivity is a crucial metric for determining the significance of parameters in model pruning \citep{movement-pruning,Super-Tickets-to-Improving-Generalization,PLATON}.  If removing a parameter (zero-out) significantly affects the loss, the model is sensitive to it.
More specifically, given a large corpus $\mathcal{D}$ and $\theta = [\theta_1, \theta_2, \ldots, \theta_d] \in \mathbb{R}^d$ as the parameters of a model, with each $\theta_j \in \mathbb{R}$ denoting the $j$-th parameter, the training objective is to minimize loss $\mathcal{L}{(\mathcal{D},\theta)}$ (defined in \ref{sec:pretrain}). The importance of each $\theta$ is denoted as $\mathcal{I}(\theta) \in \mathbb{R}^d$, where its $j$-th index $\mathcal{I}_j(\theta)$ signifies the importance for $\theta_j$. 

Under an independent and identically distributed data (i.i.d.) assumption, the importance of a parameter $\mathcal{I}_j(\theta)$ is measured by the increase in prediction loss when it is removed, calculated as the absolute difference between prediction losses with and without the parameter($\theta_j$):
\begin{equation}
    \label{eq::absolute difference loss}
    \mathcal{I}_j(\theta) = \left| \mathcal{L}{(\mathcal{D},\theta)} - \mathcal{L}{(\mathcal{D},\theta|\theta_j = 0)} \right|.
\end{equation}

Calculating $\mathcal{I}_j(\theta)$ for each parameter, as outlined in \ref{eq::absolute difference loss}, is computationally expensive because it involves $d$ distinct versions of the network computing, for each removed parameter. This becomes particularly challenging as the number of model parameters, $d$, grows to hundreds of billions. However, similar to several prior works \citep{Importance-Estimation-for-Neural-Network-Pruning,PLATON}, using the Taylor expansion formula for $\mathcal{L}$ at $\theta_j = 0$:
\begin{equation}
\begin{aligned}
    &\quad \mathcal{L}(\mathcal{D}, \theta) = \mathcal{L}(\mathcal{D}, \theta | \theta_j = 0) \\
    &+ \frac{\partial \mathcal{L}}{\partial \theta_j}(\theta_j - 0) + \frac{1}{2!} \frac{\partial^2 \mathcal{L}}{\partial \theta_j^2}(\theta_j - 0)^2 + \cdots,
\end{aligned}
\end{equation}
we can estimate $I_j(\theta)$ with its first-order Taylor expansion, eliminating the requirement for $d$ distinct networks computation:
\begin{equation}
\label{eq::gradient accumulate}
   \mathcal{I}_j(\theta) \approx \left| g_j\theta_j \right|,
\end{equation}
where $g_j = \frac{\partial \mathcal{L}}{\partial \theta_j}$ are elements of the parameter gradient $g$, and the importance is easily calculated since the gradient $g$ can be obtained from backpropagation.

\section{Experiments}
\subsection{Experimental Setup}
To localize the functional regions corresponding to linguistic competence within LLMs and analyze their nature, we perform language further pre-training (next token prediction) on various languages and observe the relationship between internal parameter removal and external output quality. We utilize LLaMA-2-7B/13B \citep{LLaMA2} as our model instance, as it stands out as one of the most notable state-of-the-art open-source LLMs in current academia. 

Our experimental dataset comprises materials from Chinese platforms like Zhihu and Wechat, English sources from Arxiv and Falcon, and a corpus including books from 28 languages, totaling 30 languages in all. Six languages, namely Arabic, Spanish, Russian, Chinese, Korean, and Vietnamese, are chosen for language further pre-training and region localization, with $100,000$ samples for each (distinct from the samples in the test set). All 30 languages are employed for model testing and functional region analysis, with the specific languages and token count detailed in \ref{sec:app_lang}. We use perplexity (PPL) as the criterion for evaluating the linguistic competence of a language model.

    \begin{table}[t]
    \centering
    \resizebox{\linewidth}{!}{
    \begin{tabular}{l cccc}
    \toprule
    \multirow{2}{*}{\textbf{Languages}} & & \multicolumn{3}{c}{LLaMA-2  \textbf{3\%} Removal}\\
    \cmidrule(r){3-5}
    & Base & Top & Bottom & Random \\
    \midrule
    \multirow{2}{*}{Arabic} & 6.771 & 127208.250 & 6.772 & 7.895 \\
    & \cellcolor{gray!20} 6.261 & \cellcolor{gray!20} 102254.758 & \cellcolor{gray!20} 6.316 & \cellcolor{gray!20} 7.112 \\
    \multirow{2}{*}{Chinese} & 8.652 & 295355.5 & 8.565 & 9.837 \\ 
    & \cellcolor{gray!20} 7.838 & \cellcolor{gray!20} 84086.906 & \cellcolor{gray!20} 7.806 & \cellcolor{gray!20} 8.619 \\
    \multirow{2}{*}{Italian} & 14.859 & 58908.879 & 14.860 & 17.341 \\
    & \cellcolor{gray!20} 13.694 & \cellcolor{gray!20} 47375.844 & \cellcolor{gray!20} 13.730 & \cellcolor{gray!20} 15.207 \\
    \multirow{2}{*}{Japanese} & 10.888 & 322031.406 & 10.896 & 12.535 \\
    & \cellcolor{gray!20} 10.072 & \cellcolor{gray!20} 75236.031 & \cellcolor{gray!20} 10.137 & \cellcolor{gray!20} 11.661 \\
    \multirow{2}{*}{Korean} & 4.965 & 125345.359 & 4.967 & 5.649 \\
    & \cellcolor{gray!20} 4.724 & \cellcolor{gray!20} 90768.844 & \cellcolor{gray!20} 4.743 & \cellcolor{gray!20} 5.241 \\
    \multirow{2}{*}{Persian} & 6.509 & 81959.719 & 6.511 & 7.628 \\
    & \cellcolor{gray!20} 6.205 & \cellcolor{gray!20} 92201.812 & \cellcolor{gray!20} 6.229 & \cellcolor{gray!20} 7.009\\
    \multirow{2}{*}{Portuguese} & 15.318 & 47763.059 & 15.319 & 17.297 \\
    & \cellcolor{gray!20} 13.667 & \cellcolor{gray!20} 51498.402 & \cellcolor{gray!20} 13.982 & \cellcolor{gray!20} 15.376 \\
    \multirow{2}{*}{Russian} & 12.062 & 170776.750 & 12.064 & 13.728 \\
    & \cellcolor{gray!20} 11.048 & \cellcolor{gray!20} 112574.609 & \cellcolor{gray!20} 10.948 & \cellcolor{gray!20} 11.757 \\
    \multirow{2}{*}{Spanish} & 17.079 & 51940.859 & 17.082 & 18.98 \\
    & \cellcolor{gray!20} 16.351 & \cellcolor{gray!20} 54005.891 & \cellcolor{gray!20} 16.138 & \cellcolor{gray!20} 17.292 \\
    \multirow{2}{*}{Ukrainian} & 9.409 & 120719.938 & 9.409 & 10.875 \\
    & \cellcolor{gray!20} 8.295 & \cellcolor{gray!20} 116287.305 & \cellcolor{gray!20} 8.297 & \cellcolor{gray!20} 9.076 \\
    \multirow{2}{*}{Vietnamese} & 5.824 & 40126.527 & 5.824 & 6.614 \\
    & \cellcolor{gray!20} 5.471 & \cellcolor{gray!20} 42336.426 & \cellcolor{gray!20} 5.437 & \cellcolor{gray!20} 5.995 \\
    \bottomrule
    \end{tabular}
    }
    \caption{LLaMA-2 perplexity on 11 languages with 3\% removal ratio. The 13B model is gray-filled while the 7B model is unfilled. `Top' and `Bottom' respectively indicate the $N$ parameters with the highest and lowest cumulative  $\mathcal{I}^*_j(\theta)$ during the further pre-training across the six languages. `Random' denotes the randomly selecting $N$ while `Base' represents no removal. Here, $N$ equals 3\% of the total number in each parameter matrix.}
    \label{tab:0.03_10W_11_scatter}
\end{table}
         
    

\subsection{Core Linguistic Competence Region}
\label{sec:localization}
In this section, we conduct further pre-training experiments on LLaMA-2 across six languages, aiming to explore and identify the core linguistic region in LLMs. Here, we define the region as "\textbf{\textit{core linguistic}}", attributed to its minimal proportion within LLMs, constituting only 1\% of the total parameters, and its association with model's linguistic competence modeling.

Specifically, according to Equation \ref{eq::gradient accumulate}, we cumulatively compute $\mathcal{I}^*(\theta)=\Sigma\mathcal{I}(\theta)$ values across six different languages' training, positing that the set of parameters exhibiting maximal importance score $\mathcal{I}^*(\theta)$ during the language further pre-training may have a strong correlation with the model's linguistic competence.
We provide both logical and empirical evidence to support this hypothesis.

\paragraph{Logical Evidence} The phenomenon of code-switching suggests that the LLMs can align languages and may possess core linguistic regions. As discussed in Section \ref{sec:importance}, if a parameter $\theta_j$ is crucial for the LLMs' core linguistic competence, the model should be sensitive to $\theta_j$, shown by a significant increase on the loss $\mathcal{L}$ when $\theta_j$ is removed, severely impairing the LLMs' linguistic performance. Conversely, other parameters impact rarely on core linguistic capabilities.

\paragraph{Empirical Evidence 1} Table \ref{tab:0.03_10W_11_scatter} illustrates that even a 3\% removal on the `Top' region leads to a substantial increase in perplexity (PPL), reaching over $40,000$ across 11 languages, indicating a complete loss of linguistic competence. In contrast, removing the `Bottom' region is comparable to non-removal `Base' in model PPL, and a `Random' removal of equal magnitude has no significant impact on the model's linguistic competence. Moreover, refer to Appendix \ref{sec:ppl_on_30}, additional experiments with reducing training samples to $10,000$ or adjusted the region selection ratio to 1\% and 5\% yield consistent findings: removing the `Top' region deprives LLaMA-2 of its capability across all 30 languages. 
This suggests the model's linguistic competence is directly influenced by the `Top' region, while removing the `Bottom' and `Random' region don't have a significant direct impact on language capabilities. See Appendix \ref{sec:ppl_on_30} for evaluations on 30 languages and further experiments.

\begin{figure*}[ht]
	\centering
	\subfigure{
		\centering
	\begin{minipage}[t]{0.31\textwidth}
        \includegraphics[width=5cm]{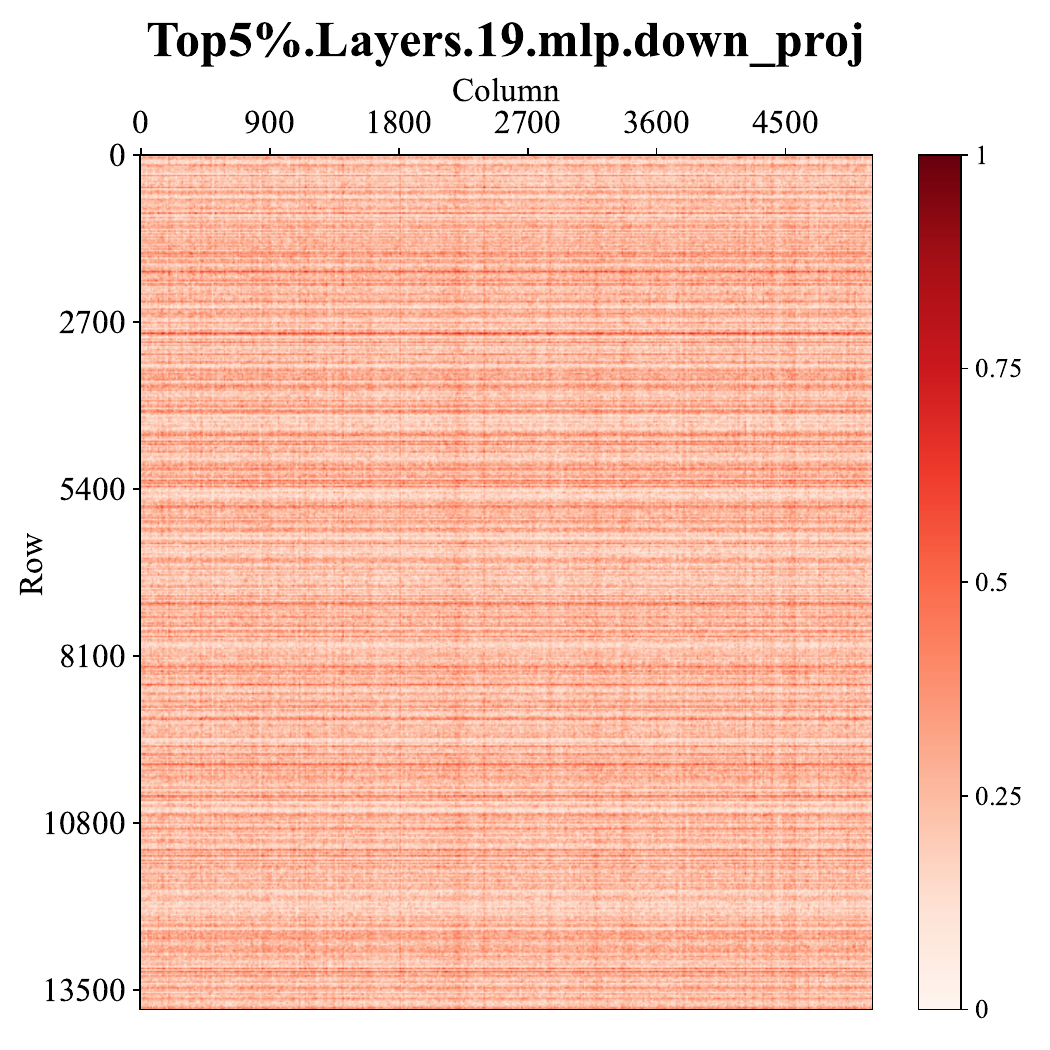}
        
	\end{minipage}
	}
	\subfigure{
		\centering
	\begin{minipage}[t]{0.31\textwidth}
            \includegraphics[width=5cm]{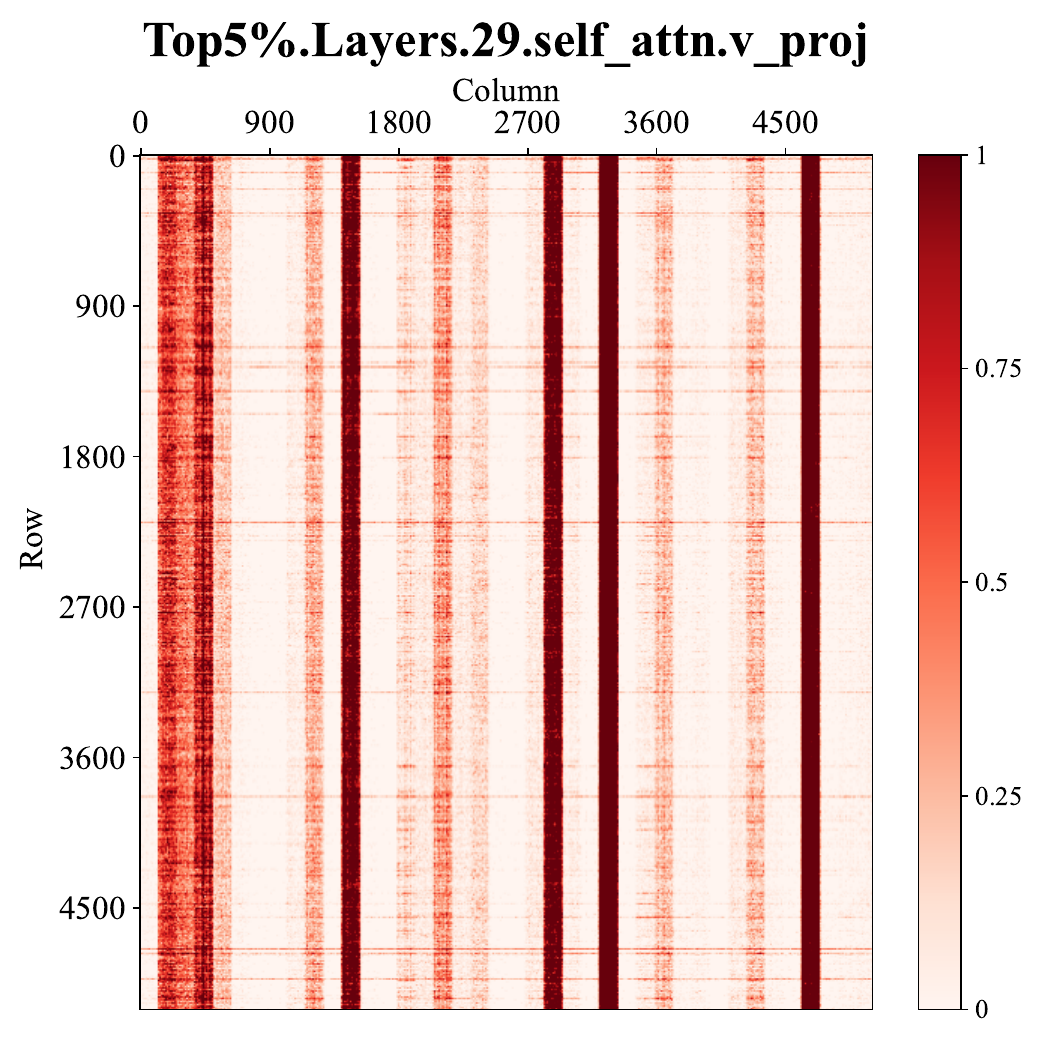}
	\end{minipage}
	}
	\subfigure{
		\centering
	\begin{minipage}[t]{0.31\textwidth}
		\includegraphics[width=5cm]{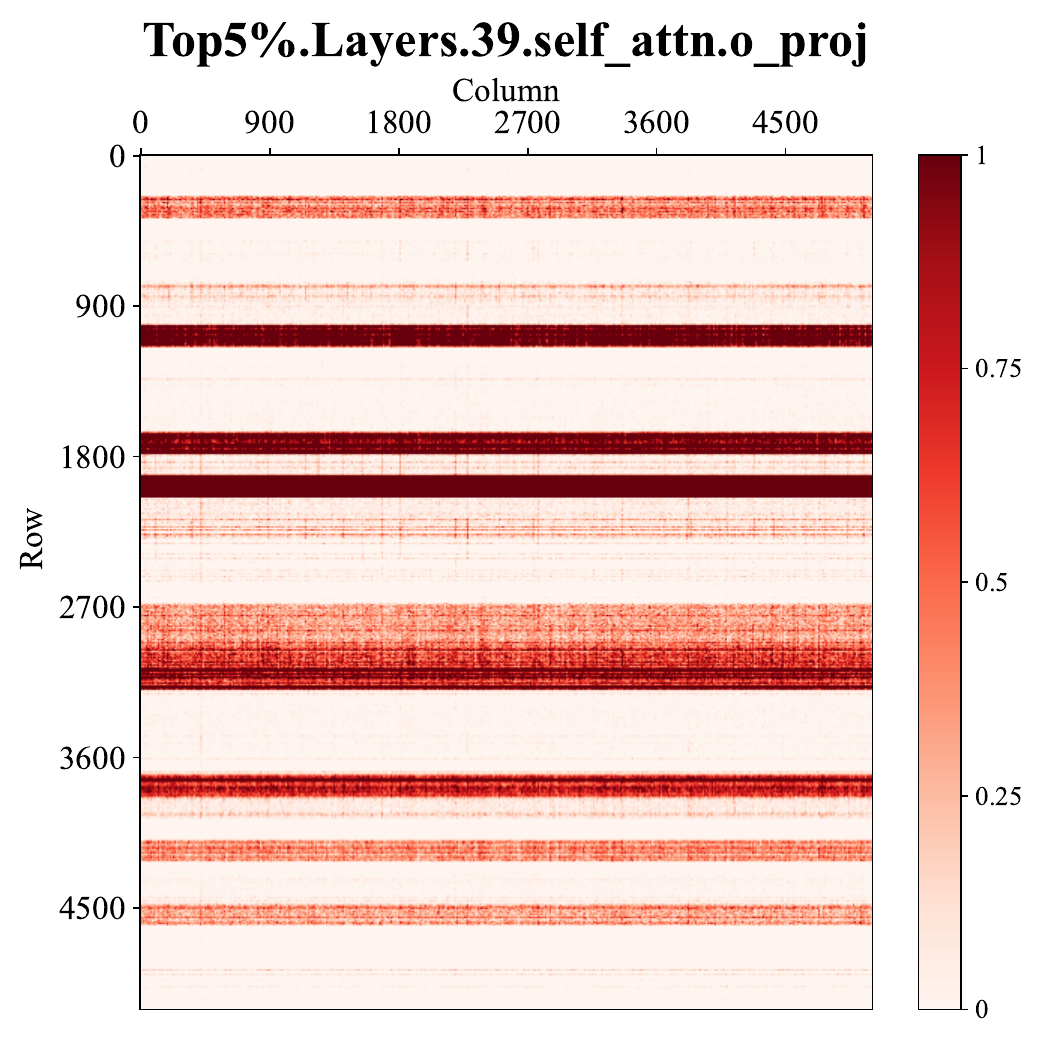}
	\end{minipage}
	}
 \caption{Visualization of the linguistic competence region (the `Top' 5\% region). The scale from 0 to 1 (after normalization) represent the proportion of parameters within a $3\times3$ vicinity that belong to the `Top' region.}
\label{fig:visualize}
\end{figure*}

\begin{table}[t]
    \centering
    \resizebox{\linewidth}{!}{
    \begin{tabular}{cc ccc}
    \toprule
    \multirow{2}{*}{\makecell[c]{Testing\\Dataset\\(Language)}} & \multirow{2}{*}{\makecell[c]{\# Training \\Samples\\ (Chinese)}} & \multicolumn{3}{c}{Removal Ratio $=1\%$} \\
    \cmidrule(r){3-5}
    & & \makecell[c]{Top \& \\Freeze} & \makecell[c]{Bottom \& \\Freeze} & \makecell[c]{Top \& \\Unfreeze}\\
    \midrule
    \multirow{7}{*}{\makecell[c]{Wechat\\(Chinese)}}
    &0K&   \cellcolor{gray!20}254772480&   6.452&   \cellcolor{gray!20}254772480\\
    &2K&  \cellcolor{gray!20}674.076&   6.052&  \cellcolor{gray!20}6.05\\
    &5K&  \cellcolor{gray!20}292.499&   6.053&  \cellcolor{gray!20}6.058\\
    &10K&  \cellcolor{gray!20}116.859&   6.305&  \cellcolor{gray!20}6.303\\
    &20K&  \cellcolor{gray!20}20.722&   6.556&  \cellcolor{gray!20}6.559\\
    &50K&  \cellcolor{gray!20}9.129&   6.18&   \cellcolor{gray!20}6.175\\
    &200K&  \cellcolor{gray!20}6.246&   5.581&   \cellcolor{gray!20}5.604\\
    \midrule
    \multirow{7}{*}{\makecell[c]{Falcon\\(English)}}
    &0K&   \cellcolor{gray!20}4244070&   14.02&   \cellcolor{gray!20}4244070\\
    &2K&  \cellcolor{gray!20}158431.282&  14.507&   \cellcolor{gray!20}14.445\\
    &5K&  \cellcolor{gray!20}343498&  15.732&  \cellcolor{gray!20}15.415\\
    &10K&  \cellcolor{gray!20}175567.219&  15.878&  \cellcolor{gray!20}15.875\\
    &20K&  \cellcolor{gray!20}32505.828&  18.689&  \cellcolor{gray!20}18.952\\
    &50K&  \cellcolor{gray!20}12455.038&  29.029&  \cellcolor{gray!20}31.583\\
    &200K&  \cellcolor{gray!20}5301.527&  488.429&   \cellcolor{gray!20}448.804\\
    \bottomrule
    \end{tabular}
    }
    \caption{Removing-freezing analysis at $1\%$ removal ratio in different regions of LLaMA-2-7B. `Top/Bottom' denotes the removal region, while `Freeze/Unfreeze' indicates whether the corresponding region is frozen after removal.}
    \label{tab:zero_and_freeze_1percent}
\end{table}

\paragraph{Empirical Evidence 2}\label{paragraph:Empirical Evidence 2} 
In the experiment corresponding to Table \ref{tab:zero_and_freeze_1percent}, we initially zero out various regions within LLaMA. Consistent with the findings from Table \ref{tab:0.03_10W_11_scatter}, removing the `Top' region leads to a loss of linguistic competence, whereas the `Bottom' region don't. However, in this experiment, we sought to ascertain if LLaMA could reacquire its lost cross-lingual generalization competence. Thus, we train on different amounts of Chinese Zhihu corpus and evaluate on Chinese Wechat and English Falcon corpora. The results indicate that unlike the `Bottom' region, if the `Top' region is removed and frozen, the model have to relearn basic language rules in other regions based on the provided Chinese Zhihu corpus, but these rules are inherently biased towards Chinese. Consequently, while its proficiency in Chinese is restored, the English perplexity remains high ($5301.527$). If the `Top' region is removed but not frozen, the model can rebuild its linguistic competence in-place. As its proficiency in Chinese is restored, so is its proficiency in English. 

This implies that the `Top' region encodes multilingual linguistic competence. When `Top' region is zeroed-out and frozen, other regions difficultly adapt to regain the core linguistic competence.

\begin{table}
    \centering
    \resizebox{\linewidth}{!}{
    \begin{tabular}{ccc cccc}
    \toprule
    \multirow{2}{*}{\makecell[c]{Model \\ Size}} & \multirow{2}{*}{\makecell[c]{\# Training \\Samples}} & \multirow{2}{*}{\makecell[c]{ $N_d$}}&\multicolumn{4}{c}{Attn.o(row), Attn.k/q/v+FFN.down(column)}\\
    \cline{4-7}
    & & &Top & Middle & Bottom & Random\\
    \midrule
    \multirow{4}{*}{\makecell[c]{7B}}
    &100K& $1$ & \cellcolor{gray!20}848.326 &  6.447 & \cellcolor{gray!20}6.447 & 6.48\\
    &100K& $3$ &\cellcolor{gray!20}72594.445 &  6.455 &  \cellcolor{gray!20}6.458 & 6.487\\
    &100K& $5$ & \cellcolor{gray!20}48001.992 &  6.461 & \cellcolor{gray!20}6.463 &  6.495\\
    &100K& $10$ & \cellcolor{gray!20}62759.516 &  6.478 &  \cellcolor{gray!20}6.48 &  6.529\\
    \hline \hline
    \multirow{4}{*}{\makecell[c]{13B}}&100K& $1$& \cellcolor{gray!20}5218.1 &  5.857 & \cellcolor{gray!20}5.857 &  5.856\\
    &100K& $3$ & \cellcolor{gray!20}37344.078 &  5.863 &  \cellcolor{gray!20}5.858 & 5.985\\
    &100K& $5$ & \cellcolor{gray!20}41840.613 &  5.867 &  \cellcolor{gray!20}5.86 & 5.89\\
    &100K& $10$ & \cellcolor{gray!20}465740.125 &  5.879 & \cellcolor{gray!20}5.869 & 6.992\\
    \hline \hline
    \multirow{4}{*}{\makecell[c]{13B}}
    &10K& $1$ & \cellcolor{gray!20}23120.977 &  5.859 &  \cellcolor{gray!20}5.856 &  5.865\\
    &10K& $3$ & \cellcolor{gray!20}28816.867 &  5.862 &  \cellcolor{gray!20}5.86 & 5.875\\
    &10K& $5$ & \cellcolor{gray!20}73268.289 &  5.866 &  \cellcolor{gray!20}5.862 & 5.878\\
    &10K& $10$ & \cellcolor{gray!20}592922.25 &  5.879 &  \cellcolor{gray!20}5.871 &  5.993\\
    \bottomrule
    \end{tabular}
    }
    \caption{Perplexity of LLaMA-2 after removing certain dimensions in the Attention and Feedforward layers. Here, $N_d$ denotes the number of dimensions to remove, `Top', `Middle', and `Bottom' refer to the dimensions with the most, moderate, and least cumulated $\mathcal{I}_\theta$ during further pre-training.
    `Random' denotes an equivalent number of dimensions chosen randomly for comparison.}
    \label{tab:att_ffn}
\end{table}

\subsection{Dimensional Dependence}
\label{sec:dependence}
To provide a more intuitive revelation of the spatial distribution characteristics of the linguistic competence region within the model, we visualize the `Top' region. As shown in Figure \ref{fig:visualize}, whether in the attention mechanism layer or the feed-forward layer, the linguistic region displays a distinct concentration in both the rows and columns of the matrices. 
In Appendix \ref{paragrah::attn.o_visualization}, we also discover that in various layers, the core linguistic region is concentrated on different heads of the Attn.o matrix.
Such distribution features seem to imply that the model's linguistic competence is localized in specific rows and columns.

\paragraph{Structured Removal}\label{paragrah::structure}
Instead of discretely removing different unstructured parameters, we selectively remove structured certain rows or columns for each matrix, especially those dimensions encompassing a significant number of `Top' region parameters, termed as `Top' dimensions. As illustrated in Table \ref{tab:att_ffn}, we attempt to remove the columns of FFN.down and Attn.k/q/v, as well as the rows of Attn.o. The results indicate that removing just these `Top' dimensions leads to a substantial decline in the model's linguistic competence. However, removals to the `Middle' ,`Bottom' and `Random' dimensions do not yield noticeable effects. Selecting the dimensional region only from the Attention matrix or inverting rows and columns removals lead to similar findings, as described in \ref{sec:attn_dimensional_removal}.

\paragraph{Single Dimension Perturbation}

\begin{figure}[t]
    \includegraphics[width=\columnwidth]{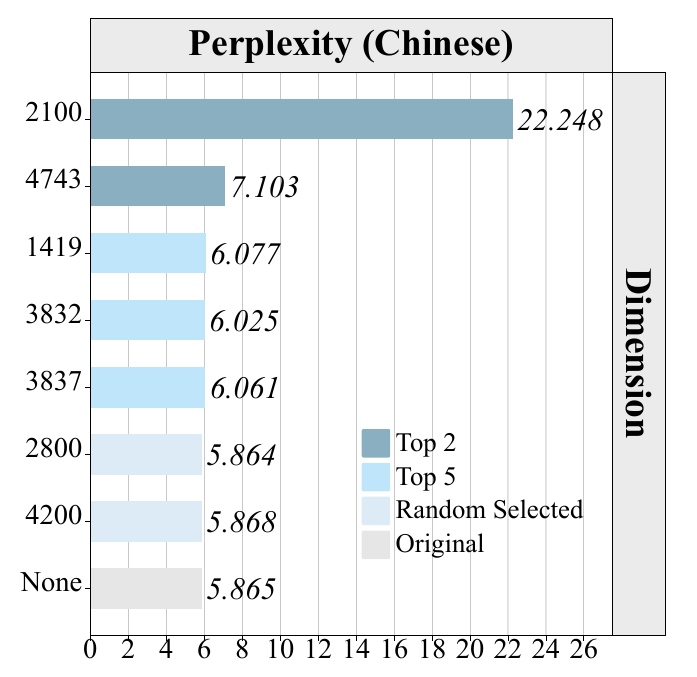}
    \caption{Perplexity of the LLaMA-2-13B when perturbing one single dimension (Att.O and FFN.down columns) across all layers.  `Top k' represents the top k dimensions that disrupt the model the most. `Random selected' refers to a randomly chosen dimension. `Original' indicates that no dimensions are disrupted.}
    \label{fig:layers}
\end{figure}

Here, we explore whether a specific dimension significantly impacts the model's linguistic competence. 
As illustrated in Figure \ref{fig:layers}, we iterate through the key dimensions mentioned in Section \ref{paragrah::structure}, attempting to perturb (random initialization) the same dimension across all Transformer layers. The results indicate that the impact of dimensions 2100 and 4743 on the LLaMA-2-13B substantially surpasses other dimensions,
even when compared to the other three in the `Top 5' dimensions. 
In contrast, perturbing two randomly selected dimensions, 2800 and 4200, yields linguistic performance almost indistinguishable from the unperturbed state. 

\paragraph{Single Parameter Perturbation}

We discover that even a slight modification to a single parameter in a model with over 13 billion parameters can lead to a significant decline in its output quality. In Table \ref{tab:norm}, merely resetting the $2100$-th parameter in the `Input\_LayerNorm' module of the $1$-st layer to its initial value causes LLaMA-2-13B's PPL value to skyrocket from 5.865 to 83224.078. If this weight parameter is multiplied by $10$, the PPL value also rises to $4363.462$. 
However, randomly altering the parameters at dimensions 2800 and 4200 doesn't noticeably impact the model. For more details, refer to Appendix \ref{sec::single_param_perturb}.

\paragraph{Ablation Study}
\label{paragraph:ablation_study}
Considering that the collapse of PPL may be possibly caused solely by the removal of outlier dimensions, rather than the collective effect of the entire linguistic region, we conduct an ablation experiment without removing the parameters of outlier dimensions (1512/2533 for LLaMA-2-7B and 2100/4743 for LLaMA-2-13B).
The results for the 13B model, presented in Table \ref{tab:ablation_outlier_dimensions_on_LLaMA-2-13B}, indicate that while outlier dimensions are contained within the core linguistic region, non-outlier dimensions in this region are also critical. Not altering any row or column of the matrix parameters for the two outlier dimensions also results in a great increase in PPL, although the collapse of PPL is slowed. This finding reveals that all dimensions in the core linguistic region are tightly interrelated, and disrupting even a small part of it can lead to a PPL collapse.
For a more detailed analysis of the experiments and results for the 7B model, please refer to Appendix \ref{sec:ablation_study}.

\begin{figure}[t]
        \includegraphics[width=\linewidth]{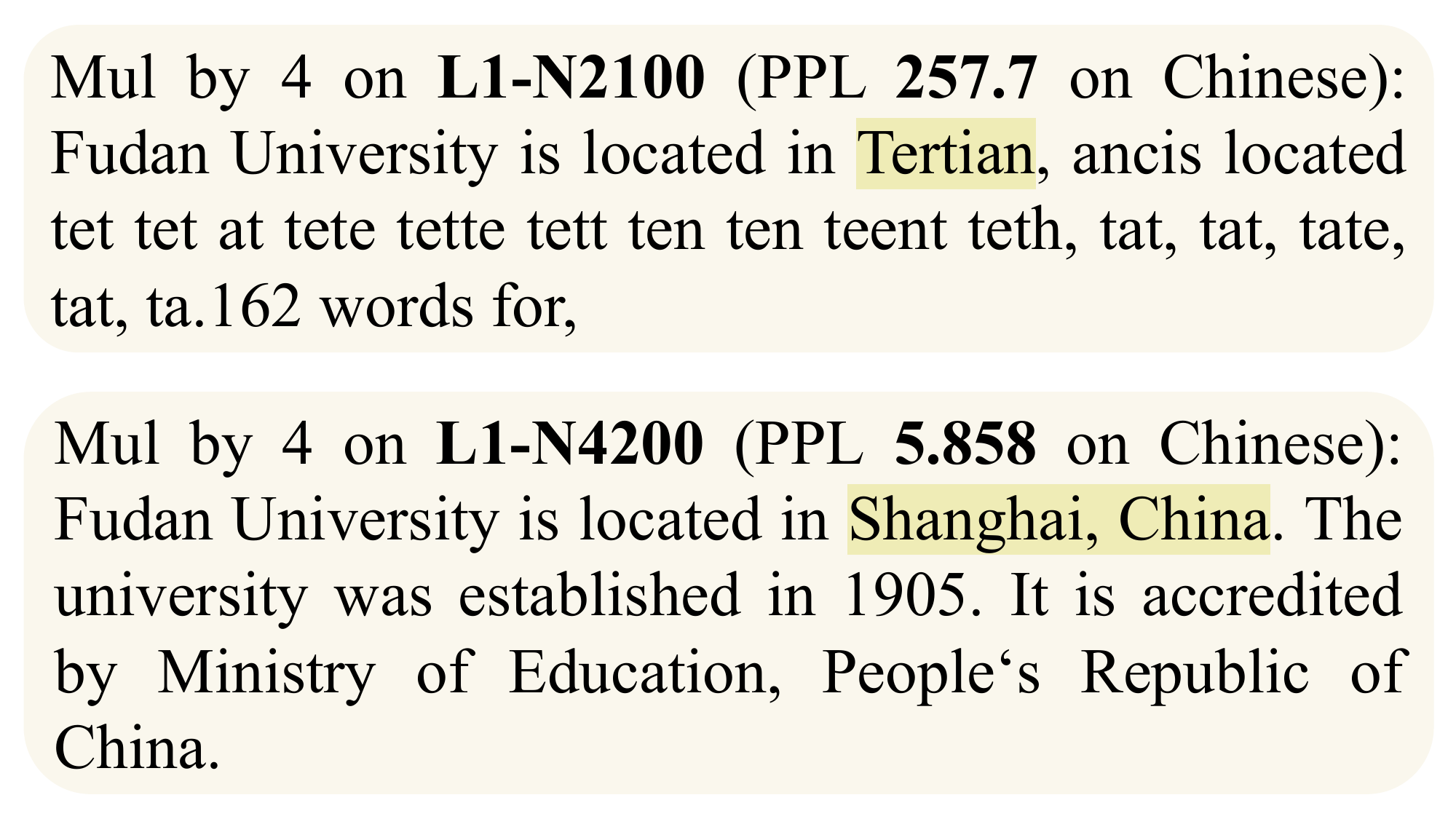}
        \caption{Comparison of linguistic competence. Expanding a single parameter to four times leads to error language competence in LLaMA-2-13B, a 13 billion-parameter LLM.}
        \label{fig:case}
    \end{figure}

\paragraph{Output Under Perturbation}To illustrate the impact of the linguistic competence region on the model's output quality, we use "\textit{Fudan University is located in}" as a prompt input and observe the model's outputs under different parameter perturbations. The results are shown in Figure \ref{fig:case}. Compared to randomly selected 4200-th dimension, perturbing model on 2100-th dimension significantly leads to model loses its linguistic competence, producing error or even nonsensical strings.

\begin{table}[t]
    \centering
    \fontsize{10pt}{11.96pt}\selectfont
    \resizebox{\linewidth}{!}{
    \begin{tabular}{c ccc}
    \toprule
    \multirow{2}{*}{\makecell[c]{ \textbf{Removal Region}}} & \multirow{2}{*}{\makecell[c]{ $N_d$}}& \multicolumn{2}{c}{LLaMA-2-\textbf{13B Top(100K)}} \\
    
    \cmidrule(r){3-4}
    & & \makecell[c]{\textbf{w/} outlier $d$} & \makecell[c]{\textbf{w/o} outlier $d$}\\
    \midrule
    \multirow{4}{*}{\makecell[c]{Attn.o(row)\\Attn.k/q/v(column)\\FFN.down(column)}}
    &1&   5218.1&   \cellcolor{gray!20}11.079\\
    &3&   37344.078&  \cellcolor{gray!20}77.519\\
    &5&   41840.613&  \cellcolor{gray!20}590.895\\
    &10&   62579.516&  \cellcolor{gray!20}513998.437\\
    \midrule
    \multirow{4}{*}{\makecell[c]{Attn.o(row)\\Attn.k/q/v(column)}}
    &1&   10.899&   \cellcolor{gray!20}10.901\\
    &3&   44.384&  \cellcolor{gray!20}44.389\\
    &5&   33.52&  \cellcolor{gray!20}29.793\\
    &10&   118.968&  \cellcolor{gray!20}120.977\\
    \midrule
    \multirow{4}{*}{\makecell[c]{Attn.o(column)\\Attn.k/q/v(row)}}
    &1&   17.609&   \cellcolor{gray!20}13.666\\
    &3&   313.178&  \cellcolor{gray!20}63.99\\
    &5&   526.464&  \cellcolor{gray!20}163.388\\
    &10&   5841.446&  \cellcolor{gray!20}2675.347\\
    \midrule
    \multirow{4}{*}{\makecell[c]{FFN.up/gate(row)\\FFN.down(column)}}
    &1&   154.925&   \cellcolor{gray!20}6.142\\
    &3&   33995.949&  \cellcolor{gray!20}6.668\\
    &5&   32572.888&  \cellcolor{gray!20}8.139\\
    &10&   524867.687&  \cellcolor{gray!20}45.408\\
    \bottomrule
    \end{tabular}
    }
    \caption{Perplexity of LLaMA-2-13B after removing `Top' certain dimensions w/ or w/o outlier dimensions respectively. Here, $N_d$ denotes the number of dimensions to remove, `Top' refers to the dimensions with the most cumulated $\mathcal{I}_\theta$ during further pre-training.}
    \label{tab:ablation_outlier_dimensions_on_LLaMA-2-13B}
\end{table}

\subsection{Monolingual Region}
\label{sec:single localization}
In this section, we wonder if LLMs possess distinct regions within different individual languages. Unlike the core linguistic regions described in Section \ref{sec:localization}, a monolingual region only has a strong correlation with certain languages, and removing it will only cause significant influence on LLMs' proficiency in those corresponding languages.

\paragraph{Region Localization}
Different from Section \ref{sec:localization}, we initially identify and select the 1\% `Top' and `Bottom' regions for each of the six languages (Arabic, Spanish, Russian, Chinese, Korean, and Vietnamese) according to Equation \ref{eq::gradient accumulate}, then deduplicate these regions. For the target language region, we exclude any regions that overlap with the `Top' and `Bottom' regions of the other five languages, aiming to eliminate the core regions and critical dimension corresponding to the model's fundamental linguistic competence. We denote $L$, $\mathcal{S}$ and $\mathcal{S}^*$ as the total set of six languages and the `Top/Bottom' regions before and after deduplication, respectively. Language $l$'s own region $\mathcal{S}_l^*$ is computed as follows:
\begin{equation}
    \mathcal{S}_l^* = \mathcal{S}_l \,\, - \bigcup_{l^{'} \in L\setminus\{l\}}\mathcal{S}_{l^{'}}.
\end{equation}
In Appendix \ref{paragraph::single_visualization}, we visualize the distribution of Attn.q matrix in `Arabic' and `Vietnamese' regions and discover minimal overlap between them.

\paragraph{Region Removal}
Unlike removing core regions or dimensions in Section \ref{sec:localization} and \ref{sec:dependence}, we discover that removing monolingual regions will only significantly affect the ability of the target languages and their closely related languages with similar letter elements or sentence structure. For example, if we remove only the region $\mathcal{S}_{Russian}^*$ for Russian alone, selected from $10,000$ or $100,000$ samples respectively, as shown in Table \ref{tab:Russian}, only Russian itself and Ukrainian have significant increases in PPL when removing `Top' $\mathcal{S}_{Russian}^*$ region. We speculate this to the fact that Russian and Ukrainian are relatively similar in terms of sentence structure and constituents, both belonging to the Slavic group. A similar phenomenon is observed if removals are changed to the regions for each of other five languages, see Appendix \ref{sec:single_language_region} for more details.

\begin{table}[t]
    \centering
    \resizebox{\linewidth}{!}{
    \begin{tabular}{l c cc cc}
    \toprule
    \multirow{2}{*}{\textbf{Languages}} & & \multicolumn{2}{c}{Russian \textbf{(10K)}} & \multicolumn{2}{c}{Russian \textbf{(100K)}}\\
    \cmidrule(r){3-4}
    \cmidrule(r){5-6}
    & Base & Top & Bottom & Top & Bottom\\
    \midrule
    Arabic & 6.771 & 7.105 & 6.785 & 7.071 & 6.787\\
    Chinese & 8.562 & 8.927 & 8.593 & 8.878 & 8.599\\ 
    Italian & 14.859 & 16.155 & 14.931 & 16.274 & 14.935\\
    Japanese & 10.888 & 11.212 & 10.931 & 11.119 & 10.951\\
    Korean & 4.965 & 5.19 & 4.972 & 5.149 & 4.974\\
    Persian & 6.509 & 6.93 & 6.506 & 6.894 & 6.515\\
    Portuguese & 15.318 & 16.51 & 15.247 & 16.421 & 15.247\\
    \cellcolor{gray!20}Russian & \cellcolor{gray!20}12.062 & \cellcolor{gray!20}\textbf{28.93} & \cellcolor{gray!20}12.141 & \cellcolor{gray!20}\textbf{41.381} & \cellcolor{gray!20}12.137\\
    Spanish & 17.079 & 18.07 & 17.224 & 17.894 & 17.211\\
    \cellcolor{gray!20}Ukrainian & \cellcolor{gray!20}9.409 & \cellcolor{gray!20}\textbf{18.147} & \cellcolor{gray!20}9.43 & \cellcolor{gray!20}\textbf{22.622} & \cellcolor{gray!20}9.435\\
    Vietnamese & 5.824 & 6.086 & 5.872 & 6.079 & 5.873\\
    \bottomrule
    \end{tabular}
    }
    \caption{LLaMA-2-7B perplexity on 11 languages with a Russian region removal. 
    Here, `Russian' and `Ukrainian' are gray-filled while others are unfilled, `Top' and `Bottom' are deduplicated, and `Base' is unchanged. Values with greater changes compared to the other regions' removals are in bold.}
    \label{tab:Russian}
\end{table}

\begin{figure}[htbp]
    \includegraphics[width=\linewidth]{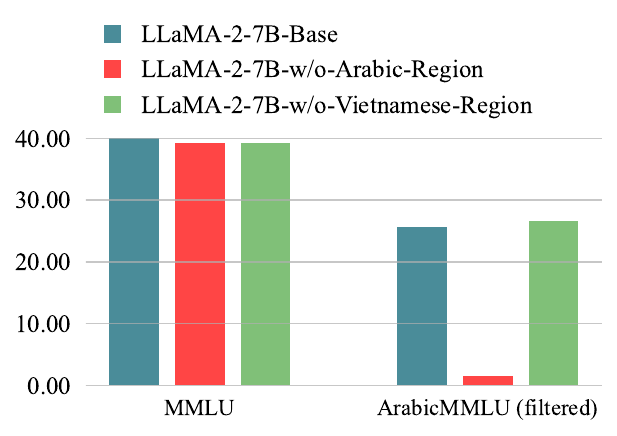}
    \caption{Model's accuracy on MMLU and ArabicMMLU (filtered) test. Here, `filtered' denotes removing questions whose correct answer is `A'.}
    \label{fig:ArabicMMLU}
\end{figure}

\paragraph{Downstream Task}
We conduct downstream tasks on MMLU \citep{MMLU} and ArabicMMLU \citep{ArabicMMLU}. The former is in English while the latter is in Arabic. Our experiments demonstrate that removing monolingual regions significantly impacts the model's ability to perform downstream tasks in the targeted language. Specifically, our findings are as follows:

\textbf{1)} The average probability for option `A' on the ArabicMMLU evaluation increases from 0.36 to 0.64 if remove the `Arabic' region.
\textbf{2)} As illustrated in Figure \ref{fig:ArabicMMLU}, the model's accuracy in the ArabicMMLU (filtered) evaluation, where the correct answer is one of the options `B/C/D/E', drops significantly from 25.6\% to merely 1.5\% when the `Arabic' region is excluded. Conversely, the removal of the `Vietnamese' region has a negligible impact on accuracy, which remains at 26.7\%.
\textbf{3)} For the MMLU evaluation, the model's accuracy is minimally impacted by the removal of monolingual regions. Compared to a baseline accuracy of 42.46\%, the removals of the `Arabic' and `Vietnamese' regions result in accuracies of 39.27\% and 39.68\%, respectively.

Furthermore, using "\textit{There are 365 days in a year and 12}" as a prompt for generation, we test outputs on the removal of the `Arabic' and `Vietnamese' regions. More details are provided in Appendix \ref{sec:case_study}.

\subsection{Further Pre-training Optimization}
In this section, we demonstrate that stabilizing the core linguistic regions (identified in Section \ref{sec:localization}) during further pre-training mitigates the catastrophic forgetting (CF) issue \citep{CF-in-Connectionist-Networks,Measuring-CF} in LLMs, while maintaining learning proficiency comparable to full-scale fine-tuning in target language. Our experimental setup involves further pre-training LLaMA-2-7B on 100,000 Arabic sentences, with a batch size of $256$, a maximum token length of $512$, and learning rates ($lr$) of $5e{-5}$ or $5e{-6}$, employing perplexity (PPL) as evaluation criterion.

\begin{figure*}[t]
	\centering
	\subfigure{
		\centering
	\begin{minipage}[t]{0.31\textwidth}
        \includegraphics[width=4.5cm]{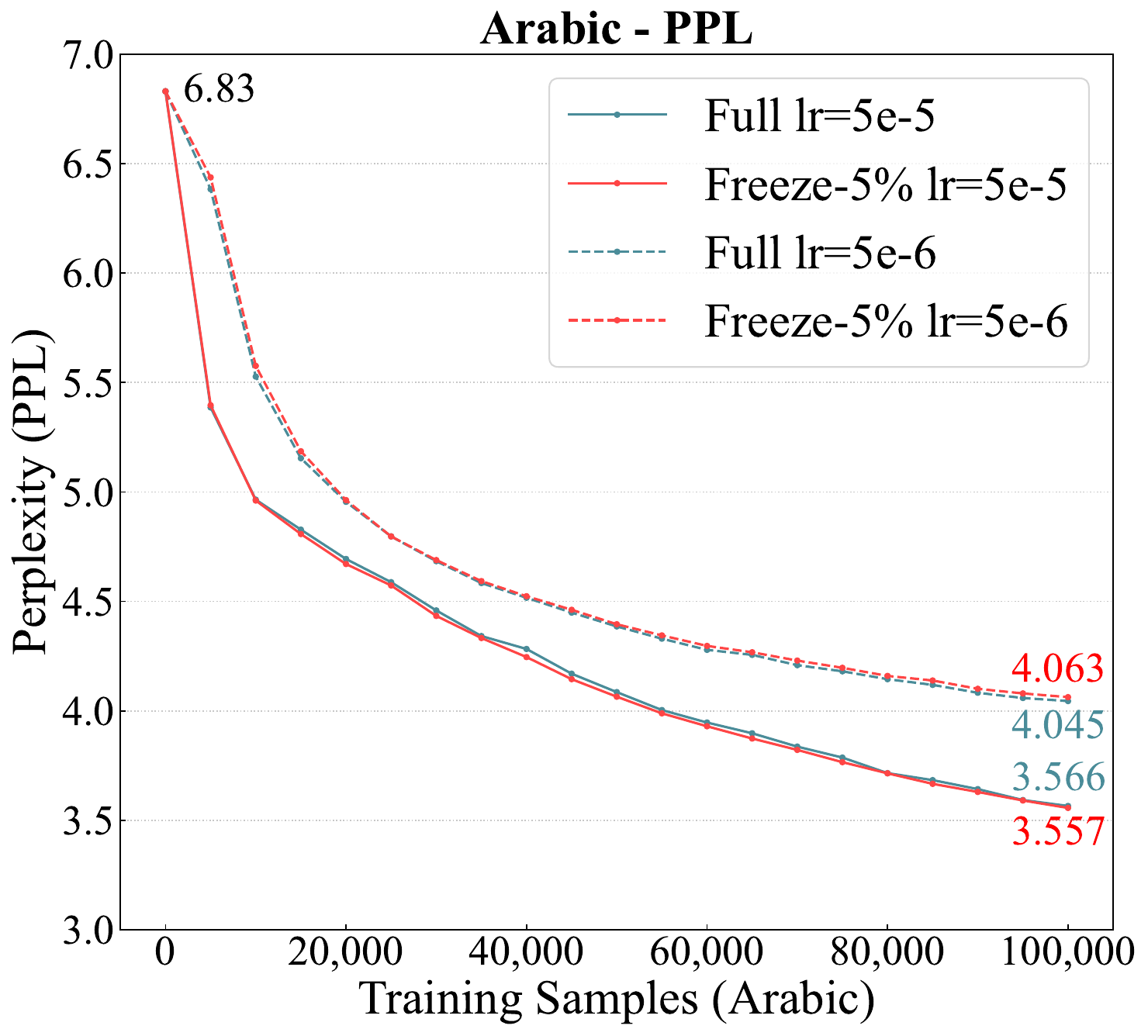}
        
	\end{minipage}
	}
	\subfigure{
		\centering
	\begin{minipage}[t]{0.31\textwidth}
            \includegraphics[width=4.5cm]{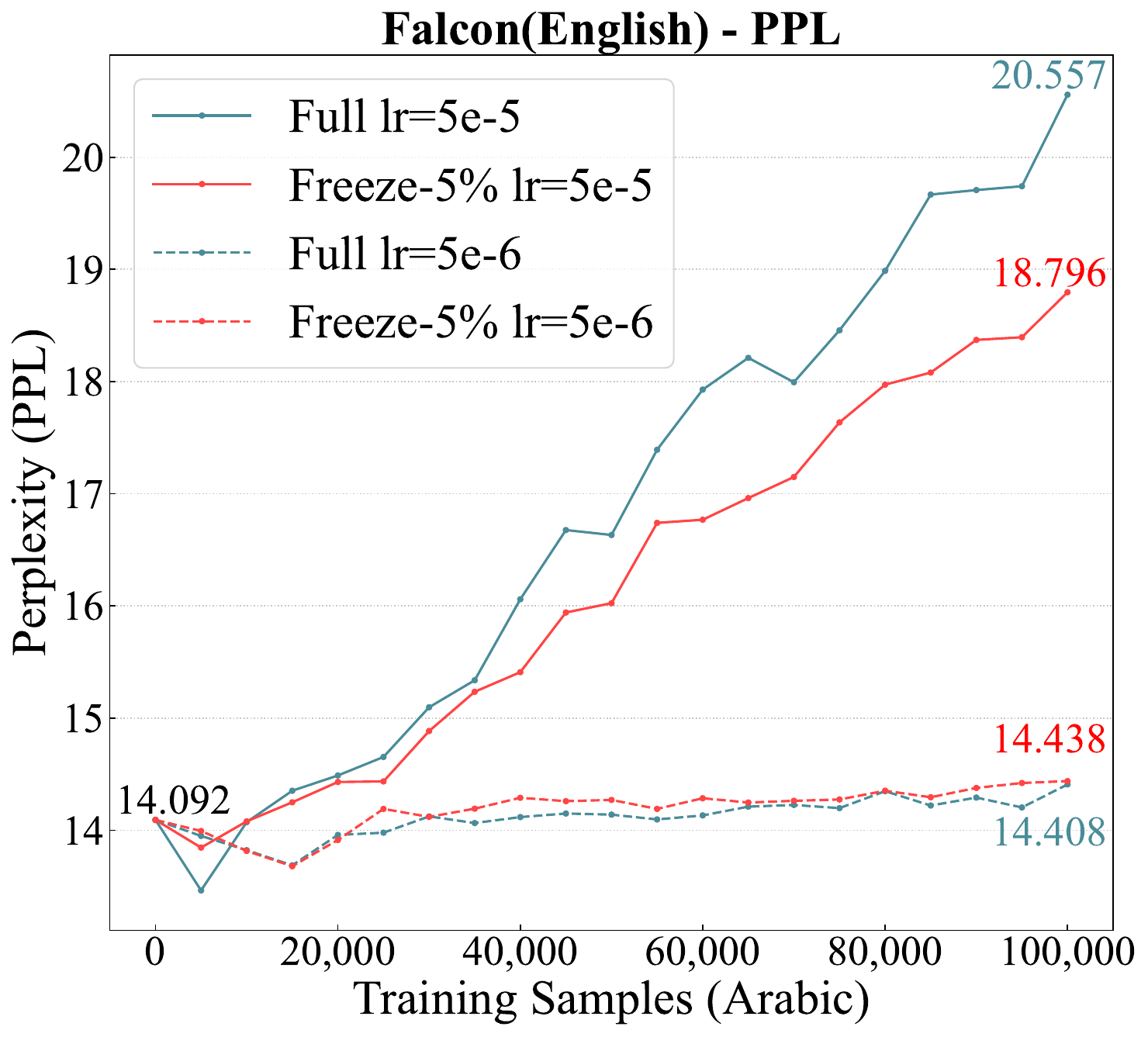}
	\end{minipage}
	}
	\subfigure{
		\centering
	\begin{minipage}[t]{0.31\textwidth}
		\includegraphics[width=5cm]{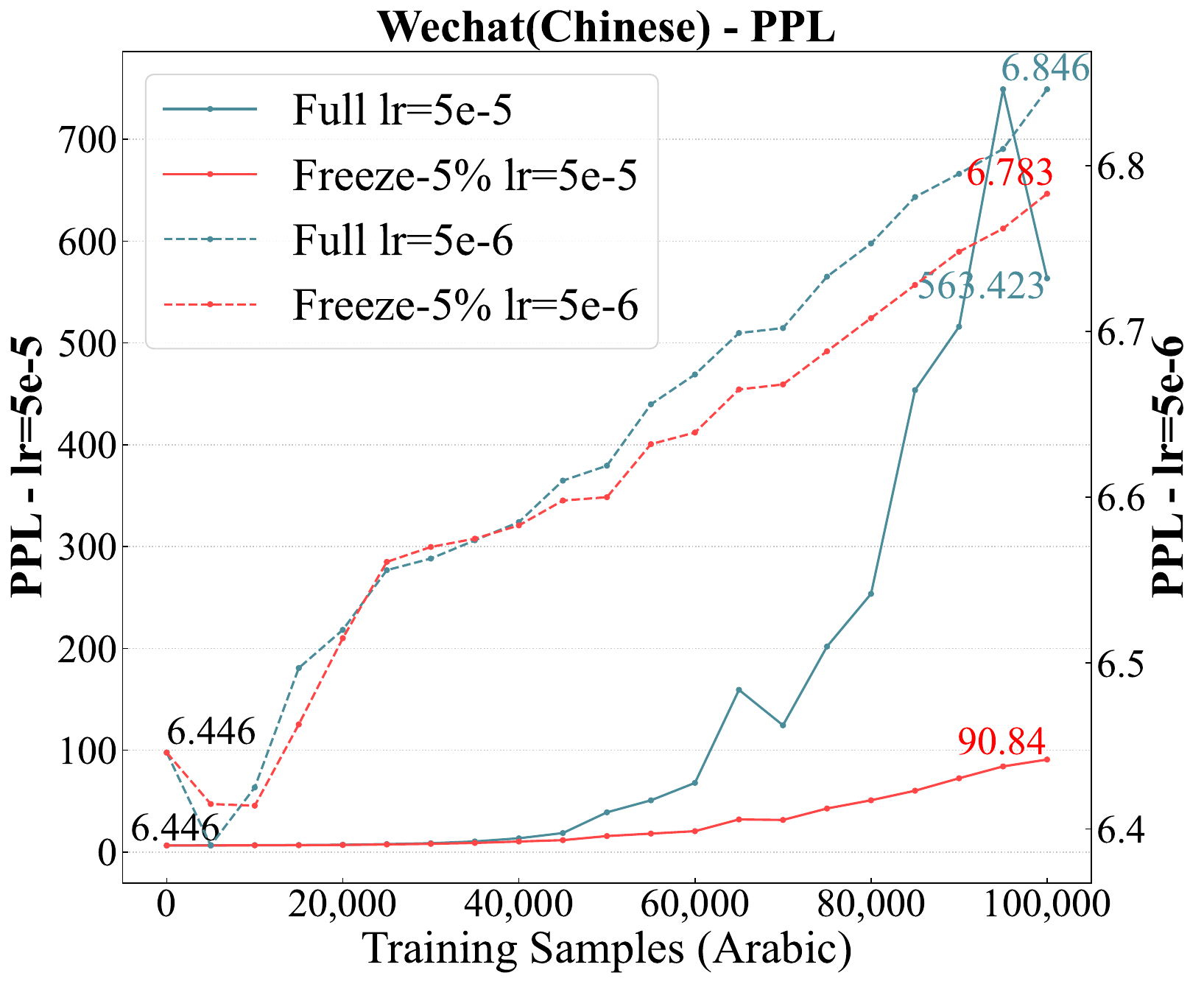}
	\end{minipage}
	}
     \caption{Perplexity of LLaMA-2 across Arabic, English, and Chinese when training on 100,000 Arabic sentences. Blue represents full-scale fine-tuning, and red denotes fine-tuning with the `Top 5\%' of the model parameters frozen. Dashed lines indicate a learning rate ($lr$) of $5e{-6}$, and solid lines represent $lr$ of $5e{-5}$. We find fine-tuning with the `Top 5\%' region frozen during further pre-training effectively mitigates forgetting of non-target languages while maintaining target language acquisition.}
    \label{fig:Finetunig-Optimization}
\end{figure*}

\paragraph{Full-scale Model Fine-tuning}Traditional full-scale fine-tuning, when increasing the learning rate or the amount of corpus data, enhances learning in the target language but aggravates forgetting in non-target languages. As shown in Table \ref{tab:zero_and_freeze_1percent}, since LLaMA-2 is primarily trained on English corpora, conducting a second stage of pre-training solely on large-scale Chinese corpora can lead to the forgetting of English competence. Additionally, as depicted on the left side of Figure \ref{fig:Finetunig-Optimization} in blue line, increasing $lr$ from $5e{-6}$ (dotted line) to $5e{-5}$ (solid line) under full-scale fine-tuning boosts the acquisition of the target language (Arabic), while simultaneously accelerates the forgetting rate of the non-target languages (English and Chinese), as shown in the middle and right side.

\paragraph{Freeze Core Regions Fine-tuning}We hypothesize that CF problem occurs due to the amplification of parameter adjustments when increasing the learning rate, which leads to significant shifts in the core linguistic region, adversely affecting language alignment. To mitigate this, we protect the core linguistic region and key dimensions by freezing the `Top 5\%' core language area for fine-tuning, as shown by the red line in Figure \ref{fig:Finetunig-Optimization}. At a $lr$ of $5e{-6}$ (dotted line), the difference between freezing fine-tuning and full-scale fine-tuning is minimal. However, when the $lr$ increases to $5e{-5}$ (solid line), freezing fine-tuning not only similarly facilitates faster learning in the target language (preserving comparable performance in Arabic PPL: 3.557 vs. 3.566), but also significantly reduces the forgetting of non-target languages (showing improvements in English and Chinese PPL: 18.796 vs. 20.557 and 90.84 vs. 563.423, respectively).

\paragraph{}
The potential reason for this phenomenon may lie in the preservation of the core regions within the cross-lingual alignment competence. 
Restricting the magnitude of updates in the core region's parameters is a future strategy that we intend to employ.
Notably, unlike regularization methods \citep{Dropout,L2regularization}, such approaches restricts to a minimal core region in LLMs, and can be implemented alongside blending previous data, retraining the entire network, or possibly only the final layers, without adding additional components to the model.

\label{sec:Further Pre-training Optimization}

\section{Related Work}
\paragraph{Intrinsic Regions}
Prior works aimed to extract a sub-network capable of executing specific downstream tasks \citep{The-Lottery-Ticket-Hypothesis, PLATON} or task-specific subspaces to limit parameter fine-tuning within it \citep{Intrinsic-Dimensionality-Explains-the-Effectiveness-of-Language-Model-Fine-Tuning, Fine-tuning-Happens-in-Tiny-Subspaces}. For parameter importance estimation, an effective metric is to use parameter magnitude \citep{To-Prune-or-Not-to-Prune, Comparing-Rewinding-and-Fine-tuning-in-Neural-Network-Pruning, Prune-Once-for-All}. 
Another metric involves estimating the sensitivity of parameters
\citep{Importance-Estimation-for-Neural-Network-Pruning, movement-pruning, Super-Tickets-to-Improving-Generalization, Importance-Estimation-with-Random-Gradient-for-Neural-Network-Pruning}.
In this work, we employ the latter method to select the most crucial parameters to unveil linguistic regions. 
Unlike \citep{The-Lottery-Ticket-Hypothesis}, which extracted a \textit{complete} sub-network for downstream tasks, as exemplified by the optimal scale of the lottery network at 21.2\%, our findings indicate that the linguistic region is a much smaller (1\%) \textit{functional} region. 

\paragraph{Cross-lingual Transfer}
Multilingual language models exhibit significant zero-shot and few-shot cross-lingual transferability across diverse tasks \citep{How-Multilingual-is-Multilingual-BERT,Are-Structural-Concepts-Universal-in-Transformer-Language-Models}. 
Fine-tuned on one language enables model to obtain comparable capabilities in another language \citep{Crosslingual-Generalization-through-Multitask-Finetuning,An-Empirical-Revisiting-on-Multilingual-Transfer-Ability}, often displaying code-switching behavior in context generation \citep{GLUECoS,LLaMA-Beyond-English}. While enhancements in cross-lingual generalization through parameter and information transfer learning \citep{UDapter,Cross-Lingual-Transfer}, compulsory language alignment \citep{Zero-Shot-Cross-lingual-Semantic-Parsing,Multilingual-Instruction-Tuning-With-Just-a-Pinch-of-Multilinguality} and in-context learning techniques \citep{few-shot-mullearning,multi-icl-alignment} have been effective, a comprehensive understanding of the internal mechanisms enabling cross-linguistic alignment in LLMs is still lacking.

\paragraph{Linguistic Abilities Probing}
Prior works have shown that multilingual LMs rely on a shared subword vocabulary and joint pre-training across multiple languages \citep{The-Surprising-Cross-Lingual-Effectiveness-of-BERT,How-Multilingual-is-Multilingual-BERT,NusaCrowd}. However, new insights highlight these models' capacity for learning universal semantic abstractions \citep{On-the-Cross-lingual-Transferability-of-Monolingual-Representations,Finding-Universal-Grammatical-Relations-in-Multilingual-BERT} and demonstrate that embeddings of similar words in similar sentences across languages are approximately aligned already \citep{Multilingual-Alignment-of-Contextual-Word-Representation,Emerging-Cross-lingual-Structure-in-Pretrained-Language-Models,Cross-Linguistic-Syntactic-Difference-in-Multilingual-mBERT}.
Analysis from a hierarchical perspective reveals that classifiers linked to different BERT \citep{BERT} layers assess semantic features through varied probe tasks \citep{Open-Sesame,What-Does-BERT-Learn-about-the-Structure-of-Language}. In this work, we introduce a parameter partitioning perspective within LLMs, identifying core linguistic and monolingual regions, which underpin cross-lingual alignment and language-specific characteristics, respectively.

\paragraph{Outlier Dimensions}
Multiple studies have identified outlier dimensions in Transformer-based LMs \citep{BERT-Busters,LLM-int8}. Researches have found that certain outlier dimensions in pre-trained LMs are highly sensitive to the fine-tuning of downstream tasks \citep{BERT-Busters,Outlier-Dimensions-that-Disrupt-Transformers-are-Driven-by-Frequency}. Furthermore, \citep{Outlier-Dimensions-that-Disrupt-Transformers-are-Driven-by-Frequency,Outlier-Dimensions-Encode-Task-Specific-Knowledge} discovered that these outlier dimensions encode task-specific knowledge, and disabling these dimensions significantly degrades model performance. \citep{Positional-Artefacts-Propagate-Through-Masked-Language-Model-Embeddings} attributed positional embeddings to the emergence of outliers, while \citep{An-Isotropy-Analysis-in-the-Multilingual-BERT-Embedding-Space} reported inconsistent results. Additionally, \citep{Massive-Activations-in-Large-Language-Models} found that these outlier dimensions exhibit significantly larger activation values than others in LLMs. Our findings demonstrate that beyond the outlier dimensions, other non-outlier dimensions within the core linguistic region also play an indispensable role in the model’s core linguistic competence.

\section{Conclusion}
This paper explores the pivotal role of certain parameters in Large Language Models (LLMs), identifying a core region essential for multilingual alignment and generalization. Removing this region causes a complete loss of linguistic competence in LLMs. Furthermore, we discover that this core region is concentrated in specific dimensions, perturbing only one dimension can cause a significant decrease in linguistic ability. Moreover, beyond the core linguistic regions, we observe that monolingual regions exist within LLMs that affect specific languages.
Importantly, we note that the catastrophic forgetting phenomenon during further pre-training may be related to drastic changes in core linguistic regions, as freezing this part during further pre-training alleviates the issue substantially. Our analysis and findings provide new perspectives and explanations for LLMs' linguistic competence.

\section*{Limitations}

In this paper, while we discover the core linguistic region and distinct monolingual regions within Large Language Models (LLMs), our work presents two notable limitations. First, our experiments are based on LLaMA-2-7B/13B, and it remains to be further determined whether the same phenomenon are observable in larger or differently architected models. Despite this, our focus on LLaMA-2-7B/13B reveals the existence of linguistic regions within the model, providing an explanation for the model's linguistic capabilities.
Secondly, we optimize full-scale fine-tuning through the freezing operation, which is not suited to extensive datasets. A more feasible approach is to limit the magnitude of parameter updates, which is the direction of our future experiments. Nevertheless, it is important to emphasize that slowing down forgetting through freezing core region suggests that in further pre-training, the core region is different from the other regions. Range of variation amplitude in core region should be smaller to maintain the cross-lingual generalization capabilities of the model. Additionally, while our study focuses on linguistic regions, beyond language, knowledge is a higher-level semantic representation, which is a critical direction for us to explore in the future.

\section*{Acknowledgements}

The authors wish to thank the anonymous reviewers for their helpful comments. This work was partially funded by National Natural Science Foundation of China (No.62076069,62206057), Shanghai Rising-Star Program (23QA1400200), and Natural Science Foundation of Shanghai (23ZR1403500).

\bibliography{acl2024}

\appendix
\section{Languages in Evaluation Corpus}
\label{sec:app_lang}
We use evaluation data composed of 30 languages to assess the model's linguistic competence. The 30 languages and their respective token counts (use LLaMA-2 Tokenizer) are as follows:
Arabic (4702998), Chinese (2869208), Czech   (1362041), Danish  (36467), Dutch (3991305), English (1216599), Finnish (372303), French  (6755281), German  (2884921), Greek   (474622), Hungarian   (1229433), Indonesian  (19226), Italian (6332560), Japanese    (501899), Korean  (2730794), Malay   (5842), Malayalam   (1489244), Norwegian   (42289), Persian (1736589), Polish  (4948702), Portuguese  (7598161), Romanian    (1381598), Russian (5205716), Spanish (7163860), Swahili (630), Swedish (1450236), Tamil   (2920808), Turkish (2484186), Ukrainian (455720), Vietnamese  (3606202).

\section{Core Linguistic Region}
\label{sec:ppl_on_30}
The regions are localized from six languages: Arabic, Spanish, Russian, Chinese, Korean, and Vietnamese, respectively. Our work does not alter the embedding layer, as we think it equates to a mapping of tokens, which does not involve modeling linguistic competence.

\begin{figure*}[htbp]
    \includegraphics[width=\linewidth]{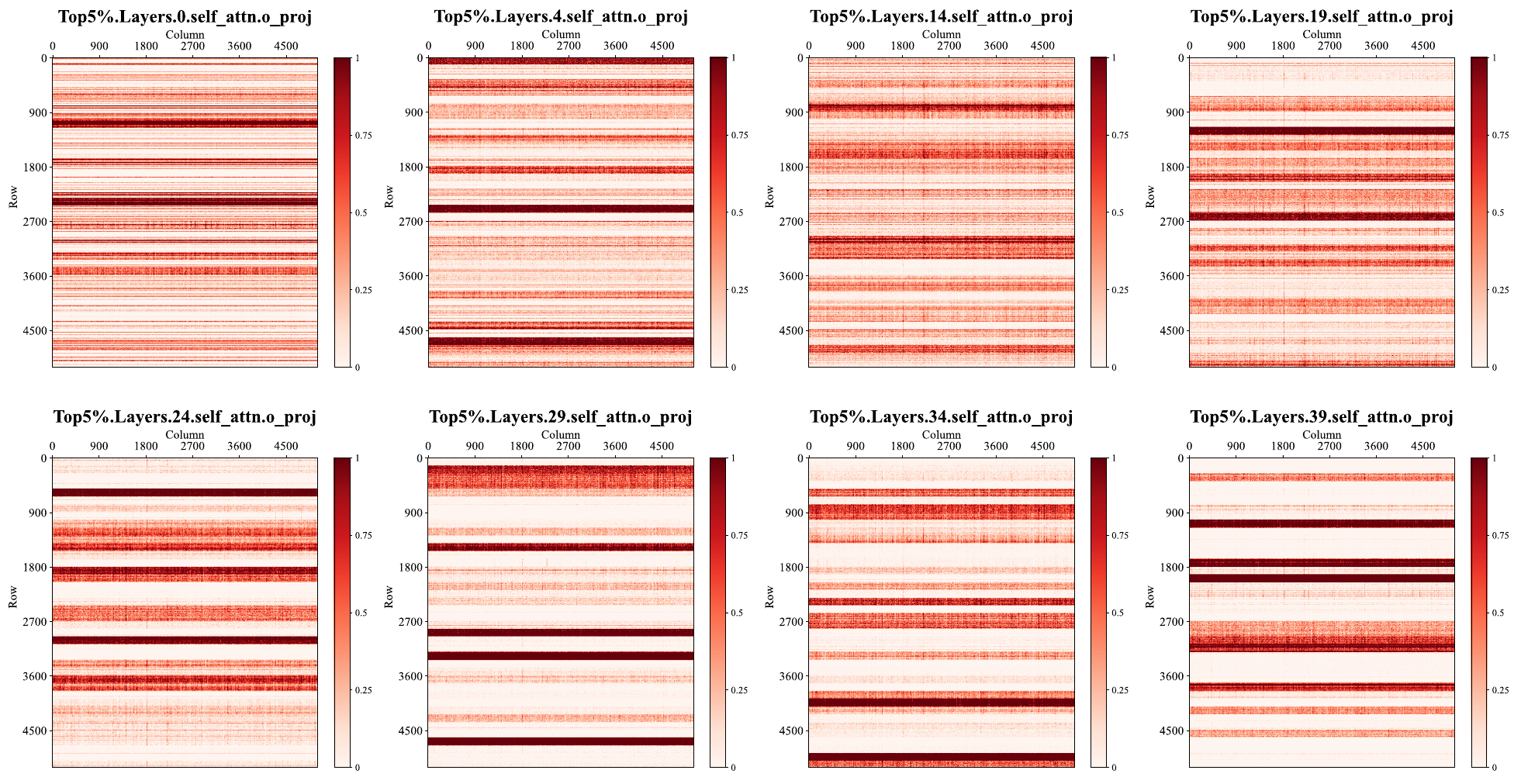}
    \caption{Visualization of the linguistic competence region (the `Top' 5\% region) in Attention.o matrix across 8 different layers. The scale from 0 to 1 (after normalization) represent the proportion of parameters within a $3\times3$ vicinity that belong to the `Top' region.}
    \label{fig:Top-Attn.o}
\end{figure*}

\paragraph{Region Visualization}\label{paragrah::attn.o_visualization} 
In Figure \ref{fig:Top-Attn.o}, we present the distribution of the `Top' 5\% regions in the Attn.o matrix for the LLaMA-2-13B model. The results indicate that across various layers, the core linguistic region on Attn.o matrix is concentrated on different rows. This difference is observed among the 40 various attention heads.

\paragraph{Removal 3\% ratio (100K)}\label{paragraph:0.03_10W_30_dev} LLaMA-2 perplexity on 30 languages when the removal ratio is 3\% ratio,  with $100,000$ samples for each language. Refer to Table \ref{tab:0.03_10W_30_scatter} for more details.

\paragraph{Removal 3\% ratio (10K)}\label{paragraph:0.03_1W_30_dev} LLaMA-2 perplexity on 30 languages when the removal ratio is 3\% ratio,  with reduced $10,000$ samples for each language. Refer to Table \ref{tab:0.03_1W_30_scatter} for more details.

\paragraph{Removal 1\% and 5\% ratio (100K)}\label{paragraph:0.01_0.05_10W_30_dev} LLaMA-2-7B perplexity on 30 languages when the removal ratio is changed to 1\% and 5\% ratio,  with $100,000$ equivalent samples for each language. Refer to Table \ref{tab:0.01_0.05_10W_30_scatter} for more details.

\section{Attention Dimensional Removal}
\label{sec:attn_dimensional_removal}
\begin{figure}[htbp]
    \includegraphics[width=\linewidth]{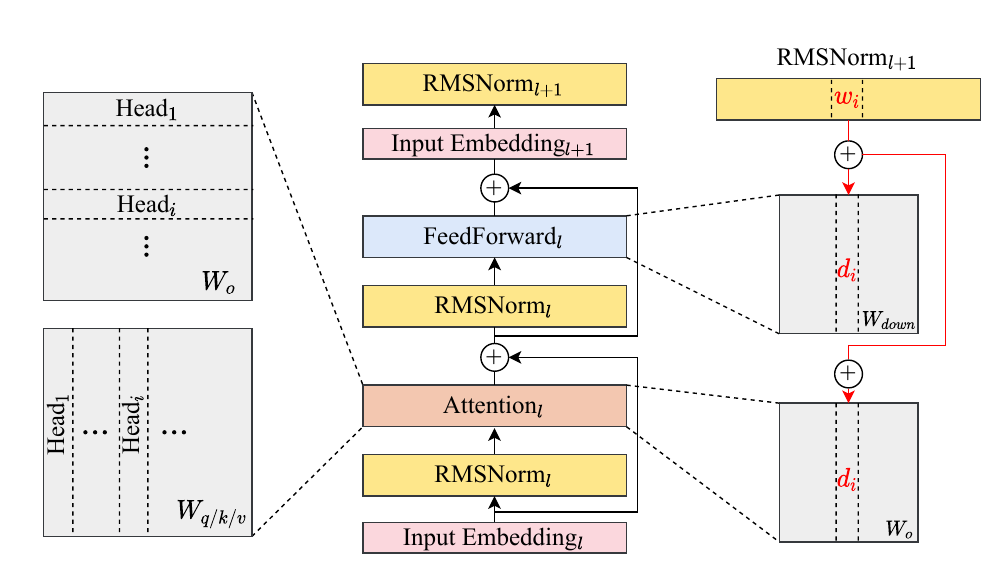}
    \caption{One can see from the left that each row of the Attn.o ($W_o$) corresponds to a particular attention head, and each column of the Attn.q/k/v ($W_{q/k/v}$) matrix corresponds to one as well. On the right, one can observe the perturbation applied to one weight within RMSNorm, which can be seen as affecting a column of the FFN.down and the Attn.o.}
    \label{fig:app_structure_visual}
\end{figure}

Figure \ref{fig:app_structure_visual} (left) illustrates that the columns of the Attn.k/q/v matrices in the attention layer, as well as the rows of the Attn.o matrix, correspond to different attention head parameters. Conversely, the rows of the Attn.k/q/v matrices and the columns of the Attn.o matrix are closely associated with dimensional features in the representation space. 

We remove the `Top' dimensions in the attention layer, and the results is displayed in Tables \ref{tab:att} and \ref{tab:att_reverse}. Table \ref{tab:att} reveals that removing the Attention layers' `Top' dimensions continues to produce more detrimental effects than other dimensions. The visualizations in Figure \ref{fig:visualize} show that these dimensions are largely concentrated in a few attention heads, suggesting that some attention heads contribute more significantly to the model's linguistic competence. Table \ref{tab:att_reverse} indicates that the removals under the second setting cause more damage than the first. Considering that, in the second setting, the `Top' dimensions in the matrix directly interact with the corresponding dimensional features in the representational space, we can conjecture that these features are tightly linked with the model's linguistic competence.

\begin{table}[htbp]
    \centering
    \resizebox{\linewidth}{!}{
    \begin{tabular}{ccc cccc}
    \toprule
    \multirow{2}{*}{\makecell[c]{Model \\ Size}} & \multirow{2}{*}{\makecell[c]{\# Training \\Samples}} & \multirow{2}{*}{\makecell[c]{ $N_d$}}&\multicolumn{4}{c}{Attn.o(row), Attn.k/q/v(column)}\\
    \cline{4-7}
    & & &Top & Middle & Bottom & Random\\
    \midrule
    \multirow{4}{*}{\makecell[c]{7B}}
    &100K& $1$ & \cellcolor{gray!20}9.731 &  6.448 & \cellcolor{gray!20}6.445 & 6.471\\
    &100K& $3$ &\cellcolor{gray!20}25.82 &  6.449 &  \cellcolor{gray!20}6.445 & 6.474\\
    &100K& $5$ & \cellcolor{gray!20}62.794 &  6.452 & \cellcolor{gray!20}6.446 &  6.482\\
    &100K& $10$ & \cellcolor{gray!20}875.016 &  6.456 &  \cellcolor{gray!20}6.446 &  6.504\\
    \hline \hline
    \multirow{4}{*}{\makecell[c]{13B}}&100K& $1$& \cellcolor{gray!20}10.899 &  5.857 & \cellcolor{gray!20}5.856 &  5.856\\
    &100K& $3$ & \cellcolor{gray!20}44.384 &  5.858 &  \cellcolor{gray!20}5.855 & 5.98\\
    &100K& $5$ & \cellcolor{gray!20}33.52 &  5.861 &  \cellcolor{gray!20}5.856 & 5.884\\
    &100K& $10$ & \cellcolor{gray!20}118.968 &  5.863 & \cellcolor{gray!20}5.857 & 5.966\\
    \hline \hline
    \multirow{4}{*}{\makecell[c]{13B}}
    &10K& $1$ & \cellcolor{gray!20}8.094 &  5.856 &  \cellcolor{gray!20}5.855 &  5.864\\
    &10K& $3$ & \cellcolor{gray!20}21.561 &  5.857 &  \cellcolor{gray!20}5.855 & 5.866\\
    &10K& $5$ & \cellcolor{gray!20}111.766 &  5.858 &  \cellcolor{gray!20}5.856 & 5.865\\
    &10K& $10$ & \cellcolor{gray!20}108.133 &  5.861 &  \cellcolor{gray!20}5.857 &  5.977\\
    \bottomrule
    \end{tabular}
    }
    \caption{Perplexity of LLaMA-2 after removing certain dimensions (zeroed-out) in the attention (Attn) layers. Here, $N_d$ denotes the number of dimensions to remove, 'Top', 'Middle', and 'Bottom' refer to the dimensions with the most, moderate, and least cumulated $\mathcal{I}_\theta$ during further pre-training across six languages, respectively. 'Random' denotes an equivalent number of dimensions chosen at random for comparison.}
    \label{tab:att}
\end{table}

\begin{table}[htbp]
    \centering
    \resizebox{\linewidth}{!}{
    \begin{tabular}{ccc cccc}
    \toprule
    \multirow{2}{*}{\makecell[c]{Model \\ Size}} & \multirow{2}{*}{\makecell[c]{\# Training \\Samples}} & \multirow{2}{*}{\makecell[c]{ $N_d$}}&\multicolumn{4}{c}{Attn.o(column), Attn.k/q/v(row)}\\
    \cline{4-7}
    & & &Top & Middle & Bottom & Random\\
    \midrule
    \multirow{4}{*}{\makecell[c]{7B}}
    &100K& $1$ & \cellcolor{gray!20}167.804 &  6.446 & \cellcolor{gray!20}6.446 & 6.446\\
    &100K& $3$ &\cellcolor{gray!20}68554.102 &  6.446 &  \cellcolor{gray!20}6.447 & 6.448\\
    &100K& $5$ & \cellcolor{gray!20}4259.861 &  6.449 & \cellcolor{gray!20}6.447 &  6.449\\
    &100K& $10$ & \cellcolor{gray!20}68170.25 &  6.454 &  \cellcolor{gray!20}6.452 &  6.449\\
    \hline \hline
    \multirow{4}{*}{\makecell[c]{13B}}&100K& $1$& \cellcolor{gray!20}17.609 &  5.855 & \cellcolor{gray!20}5.856 &  5.856\\
    &100K& $3$ & \cellcolor{gray!20}313.178 &  5.857 &  \cellcolor{gray!20}5.856 & 5.863\\
    &100K& $5$ & \cellcolor{gray!20}526.464 &  5.858 &  \cellcolor{gray!20}5.856 & 5.857\\
    &100K& $10$ & \cellcolor{gray!20}5841.446 &  5.859 & \cellcolor{gray!20}5.858 & 5.852\\
    \hline \hline
    \multirow{4}{*}{\makecell[c]{13B}}
    &10K& $1$ & \cellcolor{gray!20}17.03 &  5.855 &  \cellcolor{gray!20}5.856 &  5.857\\
    &10K& $3$ & \cellcolor{gray!20}206.225 &  5.856 &  \cellcolor{gray!20}5.856 & 5.858\\
    &10K& $5$ & \cellcolor{gray!20}1110.781 &  5.857 &  \cellcolor{gray!20}5.856 & 5.86\\
    &10K& $10$ & \cellcolor{gray!20}9600.097 &  5.859 & \cellcolor{gray!20}5.858 &  5.874\\
    \bottomrule
    \end{tabular}
    }
    \caption{Perplexity of LLaMA-2 after removing certain dimensions in attention (Attn) layers. Different from Table \ref{tab:att}, in this table, the columns of the Attn.o and the rows of the Attn.K/Q/V are removed.}
    \label{tab:att_reverse}
\end{table}

\section{Single Parameter Perturbation}
\label{sec::single_param_perturb}
In a Transformer block, each column in the Attn.o and the MLP.down matrix of the FFN layer can be considered as the input weights of a neuron. Thus, perturbing a column can be seen as disturbing the input weights of a neuron. Viewed from another angle, if we disturb the output activation value of this neuron, a similar effect should be observed. Within LLaMA, there is a specific module called RMSNorm, where each dimension is associated with a weight. Perturbations to these weights can be regarded as disturbances to the output activation values of the corresponding neurons. In Figure \ref{fig:app_structure_visual} (right), we visually demonstrate how RMSNorm affects a column of the Attn.o and the FFN.down matrix.

\begin{table}[htbp]
    \centering
    \tiny
    \fontsize{5pt}{7pt}\selectfont
    \resizebox{\linewidth}{!}{
    \begin{tabular}{ccc}
    \toprule
    Perturbation&Parameter& Perplexity\\
    \midrule
    \rowcolor{gray!20}-&- &  5.865\\
    Reset 1&L1-N2100 &  83224.078\\
    Reset 1&L1-N2800 & 5.860 \\
    Reset 1&L1-N4200 &  5.858\\
    \hline
    Mul 10&L1-N2100 &  4363.462\\
    Mul 10&L1-N2800 & 5.859 \\
    Mul 10&L1-N4200 & 5.864 \\
    \bottomrule
    \end{tabular}
    }
    \caption{Perplexity of LLaMA-2-13B on Chinese when perturbing a single weight parameter. Here, `Reset $1$' represents resetting the parameter to $1$ (the initial value before pre-training), `Mul 10' represents multiplying the parameter by $10$. `L1' represents $1$-st layers. `N' represents the `Input\_LayerNorm' module, followed by the perturbed dimension.}
    \label{tab:norm}
\end{table}

\section{Ablation Study}
\label{sec:ablation_study}
Tables \ref{tab:ablation_outlier_dimensions_on_LLaMA-2-7B} illustrate the perplexity of LLaMA-2-7B after removing core regions with and without outlier dimensions, respectively.
\paragraph{}The ablation experiments reveal that different methods of disruption and varying model sizes exhibit different rates of PPL collapse:

\textbf{1)} Removing according to Attention.Head (attn.k/q/v.col + attn.o.row) results in a slower collapse than according to Dimensional Features (attn.k/q/v.row + attn.o.col). \textbf{2)} The 13B model shows a slower rate of collapse. \textbf{3)} The abnormal dimension is mainly concentrated in the FFN layer of the “core linguistic region”. If preserving outlier dimension, the speed of PPL collapse by removing FFN layers decreases most obviously, while Attention.Head is almost unaffected.

\begin{table}[htbp]
    \centering
    \fontsize{10pt}{12.2pt}\selectfont
    \resizebox{\linewidth}{!}{
    \begin{tabular}{c ccc}
    \toprule
    \multirow{2}{*}{\makecell[c]{ \textbf{Removal Region}}} & \multirow{2}{*}{\makecell[c]{ $N_d$}}& \multicolumn{2}{c}{LLaMA-2-\textbf{7B Top(100K)}} \\
    
    \cmidrule(r){3-4}
    & & \makecell[c]{\textbf{w/} outlier d} & \makecell[c]{\textbf{w/o} outlier d}\\
    \midrule
    \multirow{4}{*}{\makecell[c]{Attn.o(row)\\Attn.k/q/v(column)\\FFN.down(column)}}
    &1&   848.326&   \cellcolor{gray!20}27.265\\
    &3&   72594.445&  \cellcolor{gray!20}57308.313\\
    &5&   48001.992&  \cellcolor{gray!20}44730.059\\
    &10&   62759.516&  \cellcolor{gray!20}73425.438\\
    \midrule
    \multirow{4}{*}{\makecell[c]{Attn.o(row)\\Attn.k/q/v(column)}}
    &1&   9.731&   \cellcolor{gray!20}9.732\\
    &3&   25.82&  \cellcolor{gray!20}25.822\\
    &5&   62.794&  \cellcolor{gray!20}23.296\\
    &10&   875.016&  \cellcolor{gray!20}860.645\\
    \midrule
    \multirow{4}{*}{\makecell[c]{Attn.o(column)\\Attn.k/q/v(row)}}
    &1&   167.804&   \cellcolor{gray!20}9.586\\
    &3&   68554.1&  \cellcolor{gray!20}136.318\\
    &5&   4259.861&  \cellcolor{gray!20}688.476\\
    &10&   68170.25&  \cellcolor{gray!20}431317.863\\
    \midrule
    \multirow{4}{*}{\makecell[c]{FFN.up/gate(row)\\FFN.down(column)}}
    &1&   20.039&   \cellcolor{gray!20}6.727\\
    &3&   74905.046&  \cellcolor{gray!20}7.672\\
    &5&   114725.578&  \cellcolor{gray!20}9.946\\
    &10&   239015.812&  \cellcolor{gray!20}16.913\\
    \bottomrule
    \end{tabular}
    }
    \caption{Perplexity of LLaMA-2-7B after removing `Top' certain dimensions w/ or w/o outlier dimensions respectively. Here, $N_d$ denotes the number of dimensions to remove, `Top' refers to the dimensions with the most cumulated $\mathcal{I}_\theta$ during further pre-training.}
    \label{tab:ablation_outlier_dimensions_on_LLaMA-2-7B}
\end{table}

\section{Monolingual Region}
\label{sec:single_language_region}
\begin{figure*}[htbp]
    \includegraphics[width=\linewidth]{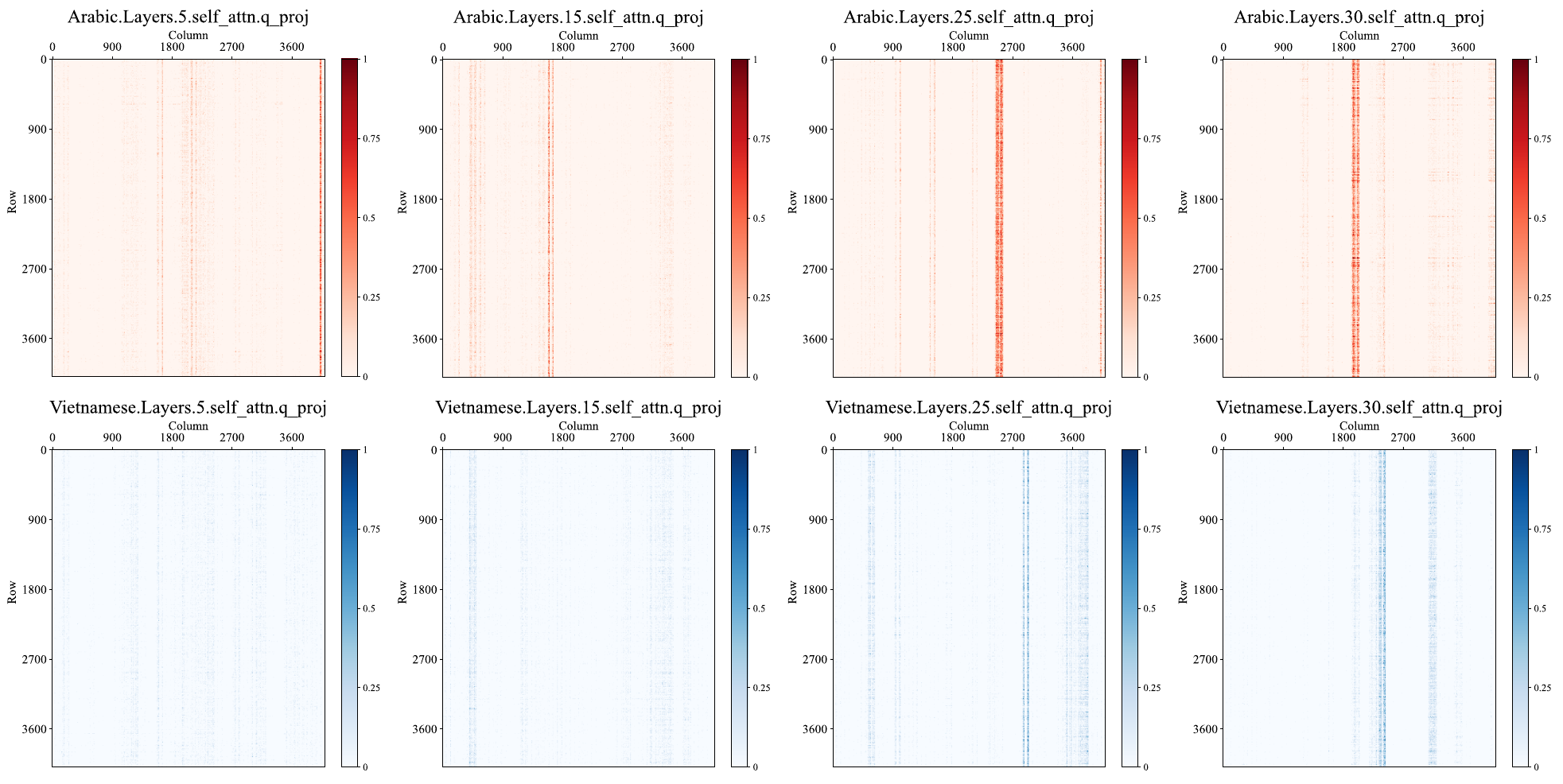}
    \caption{Visualization of the monolingual regions for `Arabic' and `Vietnamese' across 4 different layers in the Attention.q matrix. The scale from 0 to 1 (after normalization) represent the proportion of parameters within a $3\times3$ vicinity that belong to the monolingual regions.}
    \label{fig:Single-language-attn.q}
\end{figure*}

\paragraph{Region Visualization}\label{paragraph::single_visualization}
In Figure \ref{fig:Single-language-attn.q}, we present the distribution of the Attn.q matrix for `Arabic' and `Vietnamese' in 4 different layers. The results reveal that across various layers, the two monolingual regions are concentrated in different columns of the matrix.

\paragraph{Region Removal}
Tables \ref{tab:Arabic}-\ref{tab:Vietnamese} demonstrate LLaMA-2-7B perplexity after removing Arabic, Spanish, Chinese, Korean, and Vietnamese regions, respectively. The region is obtained by removing the intersections with other languages' respective regions from the 1\% `Top/Bottom' regions, selected from 10,000 or 100,000 sentences during further pre-training according to Equation \ref{eq::gradient accumulate}.

\begin{figure}[htbp]
    \includegraphics[width=\linewidth]{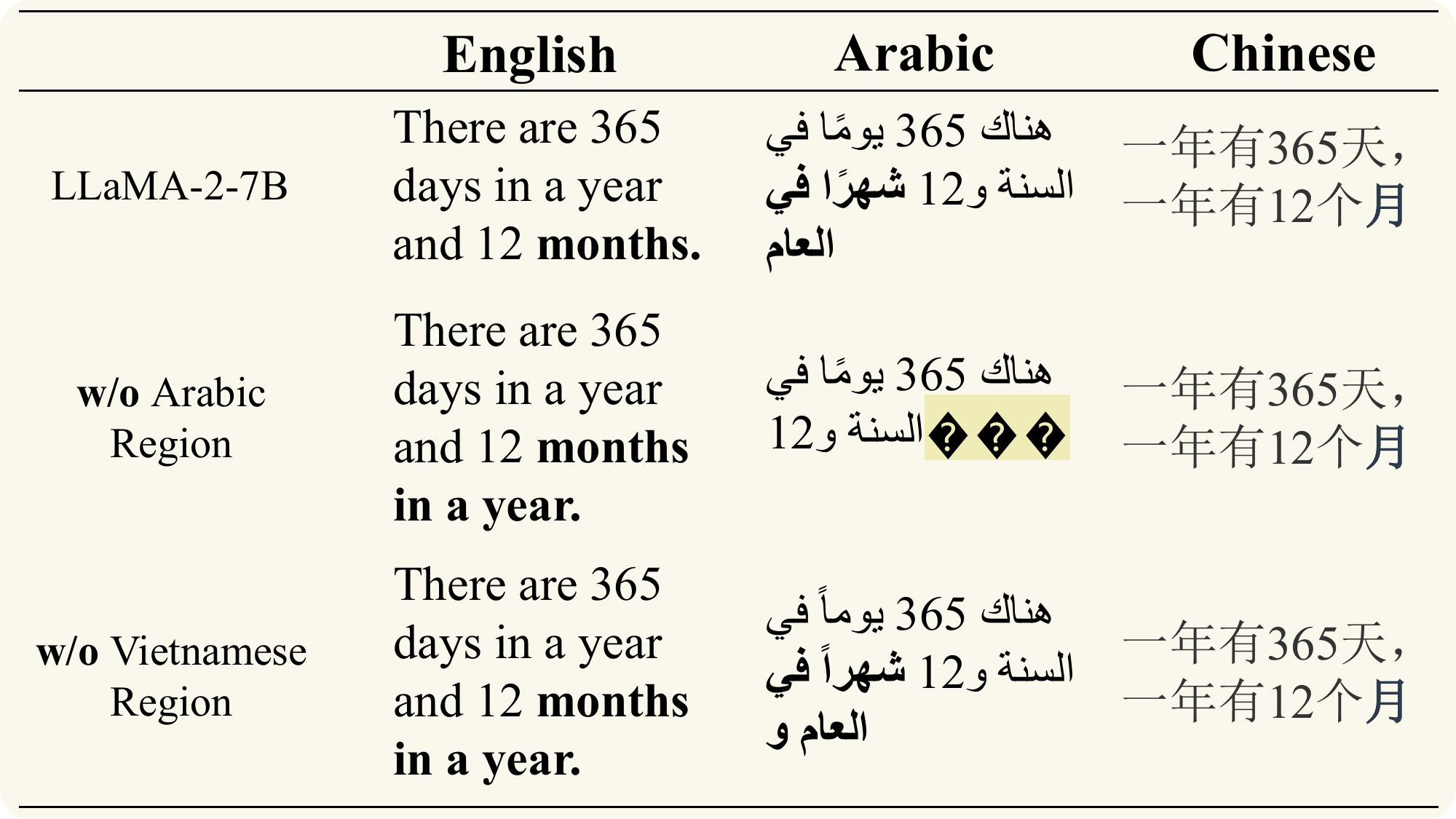}
    \caption{Model's generation with monolingual regions removed. Here, we use "\textit{There are 365 days in a year and 12}" as prompt input, and translate it into Arabic and Chinese to evaluate model's performance in three languages.}
    \label{fig:single-language-case-study}
\end{figure}

\paragraph{Case Study}
\label{sec:case_study}
In Figure \ref{fig:single-language-case-study}, we use the prompt "\textit{There are 365 days in a year and 12}" to test the model's output in English, Arabic, and Chinese, respectively. The results indicate that removing the monolingual regions causes the model to lose the relative language competence, leading the model to generate repetitive, nonsensical responses rather than correct answers like "\textit{12 months in a year}".

\begin{table}[htbp]
    \centering
    \resizebox{\linewidth}{!}{
    \begin{tabular}{l c cc cc}
    \toprule
    \multirow{2}{*}{\textbf{Languages}} & & \multicolumn{2}{c}{Arabic \textbf{(10K)}} & \multicolumn{2}{c}{Arabic \textbf{(100K)}}\\
    \cmidrule(r){3-4}
    \cmidrule(r){5-6}
    & Base & Top & Bottom & Top & Bottom\\
    \midrule
    \cellcolor{gray!20}Arabic & \cellcolor{gray!20}6.771 & \cellcolor{gray!20}\textbf{81.659} & \cellcolor{gray!20}6.785 & \cellcolor{gray!20}\textbf{135.02} & \cellcolor{gray!20}6.786\\
    Chinese & 8.562 & 9.309 & 8.593 & 9.165 & 8.588\\ 
    Italian & 14.859 & 16.61 & 14.959 & 16.366 & 14.919\\
    Japanese & 10.888 & 12.238 & 10.932 & 11.956 & 10.923\\
    Korean & 4.965 & 5.534 & 4.972 & 5.442 & 4.969\\
    \cellcolor{gray!20}Persian & \cellcolor{gray!20}6.509 & \cellcolor{gray!20}\textbf{34.142} & \cellcolor{gray!20}6.52 & \cellcolor{gray!20}\textbf{43.414} & \cellcolor{gray!20}6.508\\
    Portuguese & 15.318 & 16.909 & 15.262 & 16.86 & 15.239\\
    Russian & 12.062 & 13.708 & 12.145 & 13.781 & 12.141\\
    Spanish & 17.079 & 18.543 & 17.24 & 18.314 & 17.2\\
    Ukrainian & 9.409 & 11.243 & 9.433 & 11.225 & 9.439\\
    Vietnamese & 5.824 & 6.412 & 5.874 & 6.335 & 5.871\\
    \bottomrule
    \end{tabular}
    }
    \caption{LLaMA-2-7B perplexity on 11 languages with an `Arabic' region removal. 
    Here, `Arabic' and `Persian' are gray-filled while others are unfilled, `Top' and `Bottom' are deduplicated, and `Base' is unchanged. Values with greater changes compared to the other regions' removals are in bold.}
    \label{tab:Arabic}
\end{table}

\begin{table}[htbp]
    \centering
    \resizebox{\linewidth}{!}{
    \begin{tabular}{l c cc cc}
    \toprule
    \multirow{2}{*}{\textbf{Languages}} & & \multicolumn{2}{c}{Spanish \textbf{(10K)}} & \multicolumn{2}{c}{Spanish \textbf{(100K)}}\\
    \cmidrule(r){3-4}
    \cmidrule(r){5-6}
    & Base & Top & Bottom & Top & Bottom\\
    \midrule
    Arabic & 6.771 & 7.158 & 6.788 & 7.15 & 6.789\\
    Chinese & 8.562 & 8.984 & 8.594 & 8.971 & 8.596\\ 
     \cellcolor{gray!20}Italian &  \cellcolor{gray!20}14.859 & \cellcolor{gray!20}\textbf{21.292} & \cellcolor{gray!20}14.933 & \cellcolor{gray!20}\textbf{27.004} & \cellcolor{gray!20}14.95\\
    Japanese & 10.888 & 11.376 & 10.913 & 11.426 & 10.933\\
    Korean & 4.965 & 5.169 & 4.967 & 5.167 & 4.972\\
    Persian & 6.509 & 6.906 & 6.484 & 6.945 & 6.529\\
    \cellcolor{gray!20}Portuguese & \cellcolor{gray!20}15.318 & \cellcolor{gray!20}\textbf{21.217} & \cellcolor{gray!20}15.249 & \cellcolor{gray!20}\textbf{26.877} & \cellcolor{gray!20}15.256\\
    Russian & 12.062 & 13.039 & 12.133 & 13.252 & 12.141\\
    \cellcolor{gray!20}Spanish & \cellcolor{gray!20}17.079 & \cellcolor{gray!20}\textbf{38.876} & \cellcolor{gray!20}17.224 & \cellcolor{gray!20}\textbf{64.513} & \cellcolor{gray!20}17.225\\
    Ukrainian & 9.409 & 10.027 & 9.439 & 10.082 & 9.439\\
    Vietnamese & 5.824 & 6.136 & 5.875 & 6.145 & 5.877\\
    \bottomrule
    \end{tabular}
    }
    \caption{LLaMA-2-7B perplexity on 11 languages with a `Spanish' region removal. 
    Here, `Spanish', `Italian' and `Portuguese' are gray-filled while others are unfilled, 
    and values with greater changes compared to the other regions' removals are in bold.}
    \label{tab:Spanish}
\end{table}

\begin{table}[htbp]
    \centering
    \resizebox{\linewidth}{!}{
    \begin{tabular}{l c cc cc}
    \toprule
    \multirow{2}{*}{\textbf{Languages}} & & \multicolumn{2}{c}{Chinese \textbf{(10K)}} & \multicolumn{2}{c}{Chinese \textbf{(100K)}}\\
    \cmidrule(r){3-4}
    \cmidrule(r){5-6}
    & Base & Top & Bottom & Top & Bottom\\
    \midrule
    Arabic & 6.771 & 7.161 & 6.79 & 7.714 & 6.784\\
    \cellcolor{gray!20}Chinese & \cellcolor{gray!20}8.562 & \cellcolor{gray!20}\textbf{10.899} & \cellcolor{gray!20}8.592 & \cellcolor{gray!20}\textbf{12.079} & \cellcolor{gray!20}8.586\\ 
    Italian & 14.859 & 16.041 & 14.939 & 15.881 & 14.932\\
    \cellcolor{gray!20}Japanese & \cellcolor{gray!20}10.888 & \cellcolor{gray!20}\textbf{12.265} & \cellcolor{gray!20}10.922 & \cellcolor{gray!20}\textbf{12.878} & \cellcolor{gray!20}10.904\\
    Korean & 4.965 & 5.343 & 4.974 & 5.341 & 4.960\\
    Persian & 6.509 & 6.92 & 6.519 & 6.865 & 6.516\\
    Portuguese & 15.318 & 16.285 & 15.27 & 16.241 & 15.26\\
    Russian & 12.062 & 12.887 & 12.136 & 12.973 & 12.145\\
    Spanish & 17.079 & 18.068 & 17.216 & 17.974 & 17.219\\
    Ukrainian & 9.409 & 10.144 & 9.439 & 10.207 & 9.447\\
    Vietnamese & 5.824 & 6.261 & 5.878 & 6.296 & 5.870\\
    \bottomrule
    \end{tabular}
    }
    \caption{LLaMA-2-7B perplexity on 11 languages with a `Chinese' region removal. 
    Here, `Chinese' and `Japanese' are gray-filled while others are unfilled, 
    and values with greater changes compared to the other regions' removals are in bold.}
    \label{tab:Chinese}
\end{table}

\begin{table}[htbp]
    \centering
    \resizebox{\linewidth}{!}{
    \begin{tabular}{l c cc cc}
    \toprule
    \multirow{2}{*}{\textbf{Languages}} & & \multicolumn{2}{c}{Korean \textbf{(10K)}} & \multicolumn{2}{c}{Korean \textbf{(100K)}}\\
    \cmidrule(r){3-4}
    \cmidrule(r){5-6}
    & Base & Top & Bottom & Top & Bottom\\
    \midrule
    Arabic & 6.771 & 7.259 & 6.791 & 7.316 & 6.783\\
    Chinese & 8.562 & 9.14 & 8.594 & 9.173 & 8.594\\ 
    Italian & 14.859 & 15.91 & 14.941 & 15.791 & 14.938\\
    \cellcolor{gray!20}Japanese & \cellcolor{gray!20}10.888 & \cellcolor{gray!20}\textbf{13.273} & \cellcolor{gray!20}10.919 & \cellcolor{gray!20}\textbf{15.062} & \cellcolor{gray!20}10.932\\
    \cellcolor{gray!20}Korean & \cellcolor{gray!20}4.965 & \cellcolor{gray!20}\textbf{8.364} & \cellcolor{gray!20}4.971 & \cellcolor{gray!20}\textbf{13.128} & \cellcolor{gray!20}4.971\\
    Persian & 6.509 & 7.38 & 6.522 & 7.574 & 6.522\\
    Portuguese & 15.318 & 16.113 & 15.259 & 15.984 & 15.26\\
    Russian & 12.062 & 12.758 & 12.138 & 12.827 & 12.136\\
    Spanish & 17.079 & 17.981 & 17.214 & 17.858 & 17.225\\
    Ukrainian & 9.409 & 10.065 & 9.434 & 10.108 & 9.442\\
    Vietnamese & 5.824 & 6.188 & 5.874 & 6.177 & 5.874\\
    \bottomrule
    \end{tabular}
    }
    \caption{LLaMA-2-7B perplexity on 11 languages with a `Korean' region removal. 
    Here, `Korean' and `Japanese' are gray-filled while others are unfilled, 
    and values with greater changes compared to the other regions' removals are in bold.}
    \label{tab:Korean}
\end{table}

\begin{table}[htbp]
    \centering
    \resizebox{\linewidth}{!}{
    \begin{tabular}{l c cc cc}
    \toprule
    \multirow{2}{*}{\textbf{Languages}} & & \multicolumn{2}{c}{Vietnamese \textbf{(10K)}} & \multicolumn{2}{c}{Vietnamese \textbf{(100K)}}\\
    \cmidrule(r){3-4}
    \cmidrule(r){5-6}
    & Base & Top & Bottom & Top & Bottom\\
    \midrule
    Arabic & 6.771 & 7.435 & 6.785 & 7.341 & 6.789\\
    Chinese & 8.562 & 9.576 & 8.589 & 9.372 & 8.592\\ 
    Italian & 14.859 & 16.979 & 14.952 & 16.497 & 14.937\\
    Japanese & 10.888 & 12.027 & 10.946 & 11.814 & 10.941\\
    Korean & 4.965 & 5.44 & 4.97 & 5.335 & 4.979\\
    Persian & 6.509 & 7.315 & 6.501 & 7.243 & 6.521\\
    Portuguese & 15.318 & 17.159 & 15.249 & 16.805 & 15.258\\
    Russian & 12.062 & 13.107 & 12.141 & 13.007 & 12.144\\
    Spanish & 17.079 & 18.801 & 17.244 & 18.369 & 17.233\\
    Ukrainian & 9.409 & 10.316 & 9.447 & 10.217 & 9.433\\
    \cellcolor{gray!20}Vietnamese & \cellcolor{gray!20}5.824 & \cellcolor{gray!20}\textbf{24.382} & \cellcolor{gray!20}5.872 & \cellcolor{gray!20}\textbf{27.817} & \cellcolor{gray!20}5.874\\
    \bottomrule
    \end{tabular}
    }
    \caption{LLaMA-2-7B perplexity on 11 languages with a `Vietnamese' region removal. 
    Here, `Vietnamese' is gray-filled while others are unfilled, 
    and values with greater changes compared to the other regions' removals are in bold.}
    \label{tab:Vietnamese}
\end{table}

\begin{table*}[t]
    \centering
    \resizebox{\linewidth}{!}{
    \begin{tabular}{l cccc cccc}
    \toprule
    \multirow{2}{*}{\textbf{Languages}} & \multicolumn{4}{c}{LLaMA-2-7B \textbf{3\% (100K)}} & \multicolumn{4}{c}{LLaMA-2-13B \textbf{3\% (100K)}}\\
    \cmidrule(r){2-5}\cmidrule(r){6-9}
    & Base & Top & Bottom & Random & Base & Top & Bottom & Random \\
    \midrule
    
    Arabic & \cellcolor{gray!20} 6.771 & 127208.250 & \cellcolor{gray!20} 6.772 & 7.895 & \cellcolor{gray!20} 6.261 & 102254.758 & \cellcolor{gray!20} 6.316 & 7.112 \\
    Chinese & \cellcolor{gray!20} 8.652 & 295355.5 & \cellcolor{gray!20} 8.565 & 9.837 & \cellcolor{gray!20} 7.838 & 84086.906 & \cellcolor{gray!20} 7.806 & 8.619 \\
    Czech & \cellcolor{gray!20} 19.834 & 62692.367 & \cellcolor{gray!20} 19.835 & 24.005 & \cellcolor{gray!20} 17.744 & 56102.227 & \cellcolor{gray!20} 17.650 & 20.485 \\
    Danish & \cellcolor{gray!20} 8.372 & 47654.156 & \cellcolor{gray!20} 8.372 & 9.929 & \cellcolor{gray!20} 7.402 & 47213.586 & \cellcolor{gray!20} 7.401 & 8.278 \\
    Dutch & \cellcolor{gray!20} 16.959 & 48478.594 & \cellcolor{gray!20} 16.959 & 20.121 & \cellcolor{gray!20} 15.64 & 46303.559 & \cellcolor{gray!20} 15.572 & 18.295 \\
    English & \cellcolor{gray!20} 7.653 & 16573.422 & \cellcolor{gray!20} 7.653 & 8.359 & \cellcolor{gray!20} 7.447 & 25212.217 & \cellcolor{gray!20} 7.234 & 7.821 \\
    Finnish & \cellcolor{gray!20} 7.566 & 45711.992 & \cellcolor{gray!20} 7.566 & 8.934 & \cellcolor{gray!20} 6.887 & 48811.242 & \cellcolor{gray!20} 6.861 & 7.826 \\
    French & \cellcolor{gray!20} 13.605 & 48268.211 & \cellcolor{gray!20} 13.605 & 15.003 & \cellcolor{gray!20} 12.765 & 45674.492 & \cellcolor{gray!20} 12.573 & 13.682 \\
    German & \cellcolor{gray!20} 18.355 & 64015.117 & \cellcolor{gray!20} 18.356 & 15.404 & \cellcolor{gray!20} 17.29 & 51692.125 & \cellcolor{gray!20} 16.973 & 18.972 \\ 
    Greek & \cellcolor{gray!20} 3.832 & 224595.781 & \cellcolor{gray!20} 3.833 & 4.527 & \cellcolor{gray!20} 3.599 & 80657.891 & \cellcolor{gray!20} 3.599 & 4.146 \\
    Hungarian & \cellcolor{gray!20} 16.365 & 52828.691 & \cellcolor{gray!20} 16.363 & 20.039 & \cellcolor{gray!20} 14.756 & 58107.137 & \cellcolor{gray!20} 14.834 & 17.633 \\
    Indonesian & \cellcolor{gray!20} 44.269 & 33121.945 & \cellcolor{gray!20} 44.318 & 48.175 & \cellcolor{gray!20} 37.909 & 51611.625 & \cellcolor{gray!20} 37.838 & 38.548 \\
    Italian & \cellcolor{gray!20} 14.859 & 58908.879 & \cellcolor{gray!20} 14.860 & 17.341 & \cellcolor{gray!20} 13.694 & 47375.844 & \cellcolor{gray!20} 13.730 & 15.207 \\
    Japanese & \cellcolor{gray!20} 10.888 & 322031.406 & \cellcolor{gray!20} 10.896 & 12.535 & \cellcolor{gray!20} 10.072 & 75236.031 & \cellcolor{gray!20} 10.137 & 11.661 \\
    Korean & \cellcolor{gray!20} 4.965 & 125345.359 & \cellcolor{gray!20} 4.967 & 5.649 & \cellcolor{gray!20} 4.724 & 90768.844 & \cellcolor{gray!20} 4.743 & 5.241 \\
    Malay & \cellcolor{gray!20} 66.581 & 22603.727 & \cellcolor{gray!20} 66.843 & 74.167 & \cellcolor{gray!20} 46.885 & 40468.750 & \cellcolor{gray!20} 46.912 & 58.947 \\
    Malayalam & \cellcolor{gray!20} 5.133 & 373710.188 & \cellcolor{gray!20} 5.134 & 6.396 & \cellcolor{gray!20} 4.972 & 16990.266 & \cellcolor{gray!20} 4.972 & 5.654 \\
    Norwegian & \cellcolor{gray!20} 14.425 & 31526.176 & \cellcolor{gray!20} 14.427 & 17.854 & \cellcolor{gray!20} 13.142 & 45820.109 & \cellcolor{gray!20} 13.139 & 15.041 \\  
    Persian & \cellcolor{gray!20} 6.509 & 81959.719 & \cellcolor{gray!20} 6.511 & 7.628 & \cellcolor{gray!20} 6.205 & 92201.812 & \cellcolor{gray!20} 6.229 & 7.009\\
    Polish & \cellcolor{gray!20} 12.629 & 66906.469 & \cellcolor{gray!20} 12.629 & 14.843 & \cellcolor{gray!20} 11.414 & 55923.156 & \cellcolor{gray!20} 11.311 & 12.987 \\
    Portuguese & \cellcolor{gray!20} 15.318 & 47763.059 & \cellcolor{gray!20} 15.319 & 17.297 & \cellcolor{gray!20} 13.667 & 51498.402 & \cellcolor{gray!20} 13.982 & 15.376 \\
    Romanian & \cellcolor{gray!20} 10.893 & 43498.008 & \cellcolor{gray!20} 10.895 & 13.061 & \cellcolor{gray!20} 9.652 & 54986.055 & \cellcolor{gray!20} 9.693 & 10.969 \\
    Russian & \cellcolor{gray!20} 12.062 & 170776.750 & \cellcolor{gray!20} 12.064 & 13.728 & \cellcolor{gray!20} 11.048 & 112574.609 & \cellcolor{gray!20} 10.948 & 11.757 \\
    Spanish & \cellcolor{gray!20} 17.079 & 51940.859 & \cellcolor{gray!20} 17.082 & 18.98 & \cellcolor{gray!20} 16.351 & 54005.891 & \cellcolor{gray!20} 16.138 & 17.292 \\
    Swahili & \cellcolor{gray!20} 75.908 & 29234.168 & \cellcolor{gray!20} 75.892 & 89.380 & \cellcolor{gray!20} 70.519 & 48802.227 & \cellcolor{gray!20} 70.402 & 81.216 \\
    Swedish & \cellcolor{gray!20} 14.714 & 49425.969 & \cellcolor{gray!20} 14.714 & 17.258 & \cellcolor{gray!20} 13.229 & 48622.266 & \cellcolor{gray!20} 13.337 & 14.933 \\
    Tamil & \cellcolor{gray!20} 4.162 & 381070.844 & \cellcolor{gray!20} 4.162 & 5.04 & \cellcolor{gray!20} 4.028 & 111060.516 & \cellcolor{gray!20} 4.049 & 4.488 \\
    Turkish & \cellcolor{gray!20} 11.214 & 46986.391 & \cellcolor{gray!20} 11.215 & 13.765 & \cellcolor{gray!20} 9.834 & 50303.562 & \cellcolor{gray!20} 9.763 & 11.374 \\
    Ukrainian & \cellcolor{gray!20} 9.409 & 120719.938 & \cellcolor{gray!20} 9.409 & 10.875 & \cellcolor{gray!20} 8.295 & 116287.305 & \cellcolor{gray!20} 8.297 & 9.076 \\
    Vietnamese & \cellcolor{gray!20} 5.824 & 40126.527 & \cellcolor{gray!20} 5.824 & 6.614 &\cellcolor{gray!20}  5.471 & 42336.426 &\cellcolor{gray!20} 5.437 & 5.995 \\
    \bottomrule
    \end{tabular}
    }
    \caption{LLaMA-2 perplexity on 30 languages with 3\% removal ratio. `100K' means that the region is selected from 100,000 samples. `Top' and `Bottom' respectively indicate the $N$ parameters with the highest and lowest cumulative $\mathcal{I}^*_j(\theta)$ during the further pre-training across the six languages. `Random' denotes the randomly selecting $N$ while `Base' represents no removal. Here, $N$ equals 3\% of the total number in each parameter matrix.}
    \label{tab:0.03_10W_30_scatter}
\end{table*}

\begin{table*}[htbp]
    \centering
    \resizebox{\linewidth}{!}{
    \begin{tabular}{l cccc cccc}
    \toprule
    \multirow{2}{*}{\textbf{Languages}} & \multicolumn{4}{c}{LLaMA-2-7B  \textbf{3\% (10K)}} & \multicolumn{4}{c}{LLaMA-2-13B \textbf{3\% (10K)}}\\
    \cmidrule(r){2-5}\cmidrule(r){6-9}
    & Base & Top & Bottom & Random & Base & Top & Bottom & Random \\
    \midrule
    
    Arabic & \cellcolor{gray!20} 6.771 & 115398.328 & \cellcolor{gray!20} 6.772 & 7.895 & \cellcolor{gray!20} 6.261 & 88678.016 & \cellcolor{gray!20} 6.315 & 7.112 \\
    Chinese & \cellcolor{gray!20} 8.652 & 369027.531 & \cellcolor{gray!20} 8.564 & 9.837 & \cellcolor{gray!20} 7.838 & 70912.242 & \cellcolor{gray!20} 7.806 & 8.619 \\
    Czech & \cellcolor{gray!20} 19.834 & 78480.219 & \cellcolor{gray!20} 19.837 & 24.005 & \cellcolor{gray!20} 17.744 & 53699.43 & \cellcolor{gray!20} 17.655 & 20.485 \\
    Danish & \cellcolor{gray!20} 8.372 & 46503.742 & \cellcolor{gray!20} 8.373 & 9.929 & \cellcolor{gray!20} 7.402 & 39408.91 & \cellcolor{gray!20} 7.401 & 8.278 \\
    Dutch & \cellcolor{gray!20} 16.959 & 55704.191 & \cellcolor{gray!20} 16.961 & 20.121 & \cellcolor{gray!20} 15.64 & 41159.938 & \cellcolor{gray!20} 15.572 & 18.295 \\
    English & \cellcolor{gray!20} 7.653 & 15738.043 & \cellcolor{gray!20} 7.654 & 8.359 & \cellcolor{gray!20} 7.447 & 23678.322 & \cellcolor{gray!20} 7.234 & 7.821 \\
    Finnish & \cellcolor{gray!20} 7.566 & 46616.094 & \cellcolor{gray!20} 7.568 & 8.934 & \cellcolor{gray!20} 6.887 & 47002.539 & \cellcolor{gray!20} 6.861 & 7.826 \\
    French & \cellcolor{gray!20} 13.605 & 44385.668 & \cellcolor{gray!20} 13.609 & 15.003 & \cellcolor{gray!20} 12.765 & 38755.539 & \cellcolor{gray!20} 12.678 & 13.682 \\
    German & \cellcolor{gray!20} 18.355 & 84497.234 & \cellcolor{gray!20} 18.361 & 21.404 & \cellcolor{gray!20} 17.29 & 43319.586 & \cellcolor{gray!20} 17.02 & 18.972 \\
    Greek & \cellcolor{gray!20} 3.832 & 147740.5 & \cellcolor{gray!20} 3.833 & 4.527 & \cellcolor{gray!20} 3.599 & 70136.242 & \cellcolor{gray!20} 3.6 & 4.146 \\
    Hungarian & \cellcolor{gray!20} 16.365 & 52652.363 & \cellcolor{gray!20} 16.367 & 20.039 & \cellcolor{gray!20} 14.756 & 48407.305 & \cellcolor{gray!20} 14.735 & 17.633 \\
    Indonesian & \cellcolor{gray!20} 44.269 & 39055.945 & \cellcolor{gray!20} 44.267 & 48.175 & \cellcolor{gray!20} 37.909 & 36912.34 & \cellcolor{gray!20} 37.929 & 38.548 \\
    Italian & \cellcolor{gray!20} 14.859 & 54297.523 & \cellcolor{gray!20} 14.865 & 17.341 & \cellcolor{gray!20} 13.694 & 42515.969 & \cellcolor{gray!20} 13.69 & 15.207 \\
    Japanese & \cellcolor{gray!20} 10.888 & 358722.188 & \cellcolor{gray!20} 10.891 & 12.535 & \cellcolor{gray!20} 10.072 & 68055.984 & \cellcolor{gray!20} 10.118 & 11.661 \\
    Korean & \cellcolor{gray!20} 4.965 & 102918.828 & \cellcolor{gray!20} 4.966 & 5.649 & \cellcolor{gray!20} 4.724 & 65209.328 & \cellcolor{gray!20} 4.736 & 5.241 \\
    Malay & \cellcolor{gray!20} 66.581 & 23501.082 & \cellcolor{gray!20} 67.158 & 74.167 & \cellcolor{gray!20} 46.885 & 35517.879 & \cellcolor{gray!20} 47.191 & 58.947 \\
    Malayalam & \cellcolor{gray!20} 5.133 & 314088.969 & \cellcolor{gray!20} 5.136 & 6.396 & \cellcolor{gray!20} 4.972 & 131629.438 & \cellcolor{gray!20} 4.971 & 5.654 \\
    Norwegian & \cellcolor{gray!20} 14.425 & 38111.27 & \cellcolor{gray!20} 14.431 & 17.854 & \cellcolor{gray!20} 13.142 & 38500.664 & \cellcolor{gray!20} 13.138 & 15.041 \\
    Persian & \cellcolor{gray!20} 6.509 & 78203.031 & \cellcolor{gray!20} 6.51 & 7.628 & \cellcolor{gray!20} 6.205 & 98292.281 & \cellcolor{gray!20} 6.22 & 7.009 \\
    Polish & \cellcolor{gray!20} 12.629 & 81373.273 & \cellcolor{gray!20} 12.633 & 14.843 & \cellcolor{gray!20} 11.414 & 52403.461 & \cellcolor{gray!20} 11.393 & 12.987 \\
    Portuguese & \cellcolor{gray!20} 15.318 & 47779.789 & \cellcolor{gray!20} 15.321 & 17.297 & \cellcolor{gray!20} 13.667 & 41184.457 & \cellcolor{gray!20} 13.86 & 15.376 \\
    Romanian & \cellcolor{gray!20} 10.893 & 45836.578 & \cellcolor{gray!20} 10.897 & 13.061 & \cellcolor{gray!20} 9.652 & 51766.957 & \cellcolor{gray!20} 9.694 & 10.969 \\
    Russian & \cellcolor{gray!20} 12.062 & 227916.828 & \cellcolor{gray!20} 12.061 & 13.728 & \cellcolor{gray!20} 11.048 & 103490.719 & \cellcolor{gray!20} 11.004 & 11.757 \\
    Spanish & \cellcolor{gray!20} 17.079 & 57679.461 & \cellcolor{gray!20} 17.087 & 18.98 & \cellcolor{gray!20} 16.351 & 40338.426 & \cellcolor{gray!20} 16.265 & 17.292 \\
    Swahili & \cellcolor{gray!20} 75.908 & 42977.977 & \cellcolor{gray!20} 75.93 & 89.38 & \cellcolor{gray!20} 70.519 & 40400.949 & \cellcolor{gray!20} 70.443 & 81.216 \\
    Swedish & \cellcolor{gray!20} 14.714 & 55893.812 & \cellcolor{gray!20} 14.717 & 17.258 & \cellcolor{gray!20} 13.229 & 45396.66 & \cellcolor{gray!20} 13.301 & 14.933 \\
    Tamil & \cellcolor{gray!20} 4.162 & 447989.969 & \cellcolor{gray!20} 4.162 & 5.04 & \cellcolor{gray!20} 4.028 & 141214.188 & \cellcolor{gray!20} 4.052 & 4.488 \\
    Turkish & \cellcolor{gray!20} 11.214 & 57037.605 & \cellcolor{gray!20} 11.215 & 13.765 & \cellcolor{gray!20} 9.834 & 41566.105 & \cellcolor{gray!20} 9.791 & 11.374 \\
    Ukrainian & \cellcolor{gray!20} 9.409 & 168085.672 & \cellcolor{gray!20} 9.408 & 10.875 & \cellcolor{gray!20} 8.295 & 94307.312 & \cellcolor{gray!20} 8.296 & 9.076 \\
    Vietnamese & \cellcolor{gray!20} 5.824 & 36374.734 & \cellcolor{gray!20} 5.825 & 6.614 & \cellcolor{gray!20} 5.471 & 31730.328 & \cellcolor{gray!20} 5.467 & 5.995 \\
    
    \bottomrule
    \end{tabular}
    }
    \caption{LLaMA-2 perplexity on 30 languages with 3\% removal ratio. `10K' means that the region is selected from 10,000 samples. Here, we reduce training samples to $10,000$ during further pre-training across six languages.}
    \label{tab:0.03_1W_30_scatter}
\end{table*}

\begin{table*}[htbp]
    \centering
    \resizebox{\linewidth}{!}{
    \begin{tabular}{l cccc cccc}
    \toprule
    \multirow{2}{*}{\textbf{Languages}} & \multicolumn{4}{c}{LLaMA-2-7B  \textbf{1\% (100K)}} & \multicolumn{4}{c}{LLaMA-2-7B \textbf{5\% (100K)}}\\
    \cmidrule(r){2-5}\cmidrule(r){6-9}
    & Base & Top & Bottom & Random & Base & Top & Bottom & Random \\
    \midrule
    
    Arabic & \cellcolor{gray!20} 6.771 & 67579496 & \cellcolor{gray!20} 6.77 & 7.021 & \cellcolor{gray!20} 6.771 & 112504.609 & \cellcolor{gray!20} 6.774 & 10.823 \\
    Chinese & \cellcolor{gray!20} 8.652 & 120887480 & \cellcolor{gray!20} 8.561 & 8.818 & \cellcolor{gray!20} 8.652 & 156026.938 & \cellcolor{gray!20} 8.565 & 12.775 \\
    Czech & \cellcolor{gray!20} 19.834 & 24343856 & \cellcolor{gray!20} 19.835 & 21.176 & \cellcolor{gray!20} 19.834 & 96580.281 & \cellcolor{gray!20} 19.845 & 24.797 \\
    Danish & \cellcolor{gray!20} 8.372 & 1631186.625 & \cellcolor{gray!20} 8.372 & 8.775 & \cellcolor{gray!20} 8.372 & 82876.266 & \cellcolor{gray!20} 8.375 & 13.565 \\
    Dutch & \cellcolor{gray!20} 16.959 & 6845146 & \cellcolor{gray!20} 16.963 & 18.056 & \cellcolor{gray!20} 16.959 & 79497.211 & \cellcolor{gray!20} 16.961 & 27.01 \\
    English & \cellcolor{gray!20} 7.653 & 512756 & \cellcolor{gray!20} 7.654 & 7.851 & \cellcolor{gray!20} 7.653 & 46197.477 & \cellcolor{gray!20} 7.656 & 9.289 \\
    Finnish & \cellcolor{gray!20} 7.566 & 4727027.5 & \cellcolor{gray!20} 7.567 & 7.948 & \cellcolor{gray!20} 7.566 & 60183.328 & \cellcolor{gray!20} 7.56 & 13.005 \\
    French & \cellcolor{gray!20} 13.605 & 4768049 & \cellcolor{gray!20} 13.608 & 14.198 & \cellcolor{gray!20} 13.605 & 87642.109 & \cellcolor{gray!20} 13.611 & 19.076 \\
    German & \cellcolor{gray!20} 18.355 & 17940508 & \cellcolor{gray!20} 18.357 & 19.724 & \cellcolor{gray!20} 18.355 & 106160.992 & \cellcolor{gray!20} 18.364 & 28.772 \\
    Greek & \cellcolor{gray!20} 3.832 & 14242545 & \cellcolor{gray!20} 3.833 & 3.972 & \cellcolor{gray!20} 3.832 & 141320.578 & \cellcolor{gray!20} 3.835 & 6.45 \\
    Hungarian & \cellcolor{gray!20}16.365 & 130584 & \cellcolor{gray!20}16.366 & 17.35 & \cellcolor{gray!20}16.365 & 77265.188 & \cellcolor{gray!20}16.369 & 30.376 \\
    Indonesian & \cellcolor{gray!20}44.269 & 1654245 & \cellcolor{gray!20}44.347 & 49.476 & \cellcolor{gray!20}44.269 & 83353.344 & \cellcolor{gray!20}44.298 & 64.743 \\
    Italian & \cellcolor{gray!20}14.859 & 5265871.5 & \cellcolor{gray!20}14.863 & 15.607 & \cellcolor{gray!20}14.859 & 83076.164 & \cellcolor{gray!20}14.865 & 22.6 \\
    Japanese & \cellcolor{gray!20}10.888 & 28104000 & \cellcolor{gray!20}10.88 & 11.196 & \cellcolor{gray!20}10.888 & 124647.633 & \cellcolor{gray!20}10.895 & 16.619 \\
    Korean & \cellcolor{gray!20}4.965 & 16449047 & \cellcolor{gray!20}4.965 & 5.095 & \cellcolor{gray!20}4.965 & 59954.559 & \cellcolor{gray!20}4.967 & 7.831 \\
    Malay & \cellcolor{gray!20}66.581 & 7875206 & \cellcolor{gray!20}66.673 & 78.545 & \cellcolor{gray!20}66.581 & 51824.859 & \cellcolor{gray!20}66.751 & 90.933 \\
    Malayalam & \cellcolor{gray!20}5.133 & 7151096 & \cellcolor{gray!20}5.133 & 5.359 & \cellcolor{gray!20}5.133 & 182008.484 & \cellcolor{gray!20}5.137 & 7.905 \\
    Norwegian & \cellcolor{gray!20}14.425 & 4223085 & \cellcolor{gray!20}14.429 & 15.35 & \cellcolor{gray!20}14.425 & 79399.109 & \cellcolor{gray!20}14.434 & 23.621 \\
    Persian & \cellcolor{gray!20}6.509 & 2233196 & \cellcolor{gray!20}6.507 & 6.782 & \cellcolor{gray!20}6.509 & 107342.734 & \cellcolor{gray!20}6.511 & 10.236 \\
    Polish & \cellcolor{gray!20}12.629 & 6547834.5 & \cellcolor{gray!20}12.631 & 13.36 & \cellcolor{gray!20}12.629 & 88912.945 & \cellcolor{gray!20}12.632 & 21.372 \\
    Portuguese & \cellcolor{gray!20}15.318 & 6249820 & \cellcolor{gray!20}15.319 & 15.927 & \cellcolor{gray!20}15.318 & 78851.766 & \cellcolor{gray!20}15.324 & 22.608 \\
    Romanian & \cellcolor{gray!20} 10.893 & 5251915.5 & \cellcolor{gray!20} 10.895 & 11.526 & \cellcolor{gray!20} 10.893 & 71228.375 & \cellcolor{gray!20} 10.899 & 19.21 \\
    Russian & \cellcolor{gray!20} 12.062 & 17596800 & \cellcolor{gray!20} 12.061 & 12.067 & \cellcolor{gray!20} 12.062 & 102639.602 & \cellcolor{gray!20} 12.066 & 18.504 \\
    Spanish & \cellcolor{gray!20} 17.079 & 8220832.5 & \cellcolor{gray!20} 17.084 & 18.029 & \cellcolor{gray!20} 17.079 & 96575.547 & \cellcolor{gray!20} 17.084 & 24.007 \\
    Swahili & \cellcolor{gray!20} 75.908 & 7875009 & \cellcolor{gray!20} 75.845 & 83.963 & \cellcolor{gray!20} 75.908 & 77765.133 & \cellcolor{gray!20} 75.8 & 131.709 \\
    Swedish & \cellcolor{gray!20} 14.714 & 4712167.5 & \cellcolor{gray!20} 14.716 & 15.534 & \cellcolor{gray!20} 14.714 & 81574.734 & \cellcolor{gray!20} 14.717 & 22.628 \\
    Tamil & \cellcolor{gray!20} 4.162 & 20660974 & \cellcolor{gray!20} 4.162 & 4.265 & \cellcolor{gray!20} 4.162 & 173728.312 & \cellcolor{gray!20} 4.164 & 5.881 \\
    Turkish & \cellcolor{gray!20} 11.214 & 4489915 & \cellcolor{gray!20} 11.214 & 11.882 & \cellcolor{gray!20} 11.214 & 58347.055 & \cellcolor{gray!20} 11.218 & 19.76 \\
    Ukrainian & \cellcolor{gray!20} 9.409 & 11689088 & \cellcolor{gray!20} 9.409 & 9.811 & \cellcolor{gray!20} 9.409 & 90008.312 & \cellcolor{gray!20} 9.414 & 14.807 \\
    Vietnamese & \cellcolor{gray!20} 5.824 & 2235468 & \cellcolor{gray!20} 5.825 & 6.018 & \cellcolor{gray!20} 5.824 & 54187.02 & \cellcolor{gray!20} 5.825 & 9.015 \\
    \bottomrule
    \end{tabular}
    }
    \caption{LLaMA-2-7B perplexity on 30 languages with 1\% and 5\% removal ratio. `100K' means that the region is selected from 100,000 samples. Here, we change the removal ratio from 3\% to 1\% and 5\%.}
    \label{tab:0.01_0.05_10W_30_scatter}
\end{table*}

\end{document}